\documentclass[sigconf,usenames,dvipsnames]{acmart}

\usepackage{balance}
\usepackage{booktabs} \usepackage{amsmath}
\usepackage{amsfonts}
\usepackage{amssymb}
\usepackage{bm}
\usepackage{graphicx}
\usepackage{paralist}
\usepackage{enumitem}
\setlist[itemize]{nosep,leftmargin=2em}  \setlist[enumerate]{nosep,leftmargin=2em}  \usepackage{xspace}
\usepackage[labelformat=simple]{subcaption}

\usepackage{makecell}
\usepackage{hyperref}
\usepackage{cleveref}
\usepackage[vlined,ruled,linesnumbered]{algorithm2e}
\usepackage{multirow}
\usepackage{kotex}
\usepackage[breakwords]{truncate}

\def\BibTeX{{\rm B\kern-.05em{\sc i\kern-.025em b}\kern-.08emT\kern-.1667em\lower.7ex\hbox{E}\kern-.125emX}}

\copyrightyear{2019} 
\acmYear{2019} 
\setcopyright{acmcopyright}
\acmConference[KDD '19]{The 25th ACM SIGKDD Conference on Knowledge Discovery and Data Mining}{August 4--8, 2019}{Anchorage, AK, USA}
\acmBooktitle{The 25th ACM SIGKDD Conference on Knowledge Discovery and Data Mining (KDD '19), August 4--8, 2019, Anchorage, AK, USA}
\acmPrice{15.00}
\acmDOI{10.1145/3292500.3330855}
\acmISBN{978-1-4503-6201-6/19/08}

\renewcommand{\thefootnote}{\fnsymbol{footnote}}
\newcommand{\eat}[1]{}

\newcommand{\tablefont}{}

\newcommand{\mat}[1]{\mathbf{#1}}

\newcommand{\N}{{\mathcal{N}}}
\newcommand{\nodeWithScores}{V_s}

\newcommand{\tmdb}{\textsc{tmdb5k}\xspace}
\newcommand{\music}{\textsc{music10k}\xspace}
\newcommand{\fb}{\textsc{fb15k}\xspace}
\newcommand{\imdb}{\textsc{imdb}\xspace}

\newcommand{\method}{\textsc{GENI}\xspace}
\newcommand{\pr}{\textsc{PR}\xspace}
\newcommand{\ppr}{\textsc{PPR}\xspace}
\newcommand{\rwr}{\textsc{RWR}\xspace}
\newcommand{\har}{\textsc{HAR}\xspace}
\newcommand{\logindeg}{\textsc{LID}\xspace}
\newcommand{\linreg}{\textsc{LR}\xspace}
\newcommand{\randforest}{\textsc{RF}\xspace}
\newcommand{\neuralnet}{\textsc{NN}\xspace}
\newcommand{\fcnn}{\textsc{FCNN}\xspace}
\newcommand{\gat}{\textsc{GAT}\xspace}

\newcommand{\spearman}{\textsc{Spearman}\xspace}
\newcommand{\ndcg}{\textsc{NDCG@100}\xspace}
\newcommand{\ndcgt}{\textsc{NDCG@2000}\xspace}

\begin{document}

\title[Estimating Node Importance in Knowledge Graphs Using Graph Neural Networks]{Estimating Node Importance in Knowledge Graphs Using\\ Graph Neural Networks}

\author{Namyong Park$^{1*}$, Andrey Kan$^2$, Xin Luna Dong$^2$, Tong Zhao$^2$, Christos Faloutsos$^{1*}$}
\thanks{${}^*$Work performed while at Amazon.}
\email{{namyongp,christos}@cs.cmu.edu,{avkan,lunadong,zhaoton}@amazon.com}
\affiliation{\institution{\textsuperscript{1}Carnegie Mellon University, \textsuperscript{2}Amazon}}

\renewcommand{\shortauthors}{N. Park et al.}

\begin{abstract}
How can we estimate the importance of nodes in a knowledge graph (KG)?
A KG is a multi-relational graph that has proven valuable for many tasks including
question answering and semantic search. 
In this paper, we present \method, a method for tackling the problem of estimating node importance in KGs,
which enables several downstream applications such as item recommendation and resource allocation. 
While a number of approaches have been developed to address this problem for general graphs,
they do not fully utilize information available in KGs, or
lack flexibility needed to model complex relationship between entities and their importance.
To address these limitations, we explore supervised machine learning algorithms.
In particular, building upon recent advancement of graph neural networks (GNNs),
we develop \method, a GNN-based method designed to deal with distinctive challenges 
involved with predicting node importance in KGs.
Our method performs an aggregation of importance scores
instead of aggregating node embeddings
via predicate-aware attention mechanism and flexible centrality adjustment.
In our evaluation of \method and existing methods 
on predicting node importance in real-world KGs with different characteristics,
\method achieves 5--17\% higher NDCG@100 than the state of the art.
\end{abstract}



\maketitle

\renewcommand*{\thefootnote}{\arabic{footnote}}
\setcounter{footnote}{0}

\abovedisplayskip=2.0pt  
\belowdisplayskip=2.0pt

\captionsetup[table]{skip=2pt}  

{\fontsize{8pt}{8pt} \selectfont
\textbf{ACM Reference Format:}\\
Namyong Park, Andrey Kan, Xin Luna Dong, Tong Zhao, Christos Faloutsos. 2019. Estimating Node Importance in Knowledge Graphs Using Graph Neural Networks. In \textit{The 25th ACM SIGKDD Conference on Knowledge Discovery and Data Mining (KDD '19), August 4--8, 2019, Anchorage, AK, USA.} ACM, New York, NY, USA, 11 pages. https://doi.org/10.1145/3292500.3330855}

\section{Introduction}
\label{sec:intro}
\begin{figure}[!t]
\centering
\makebox[\linewidth][c]{\includegraphics[width=1.1\linewidth]{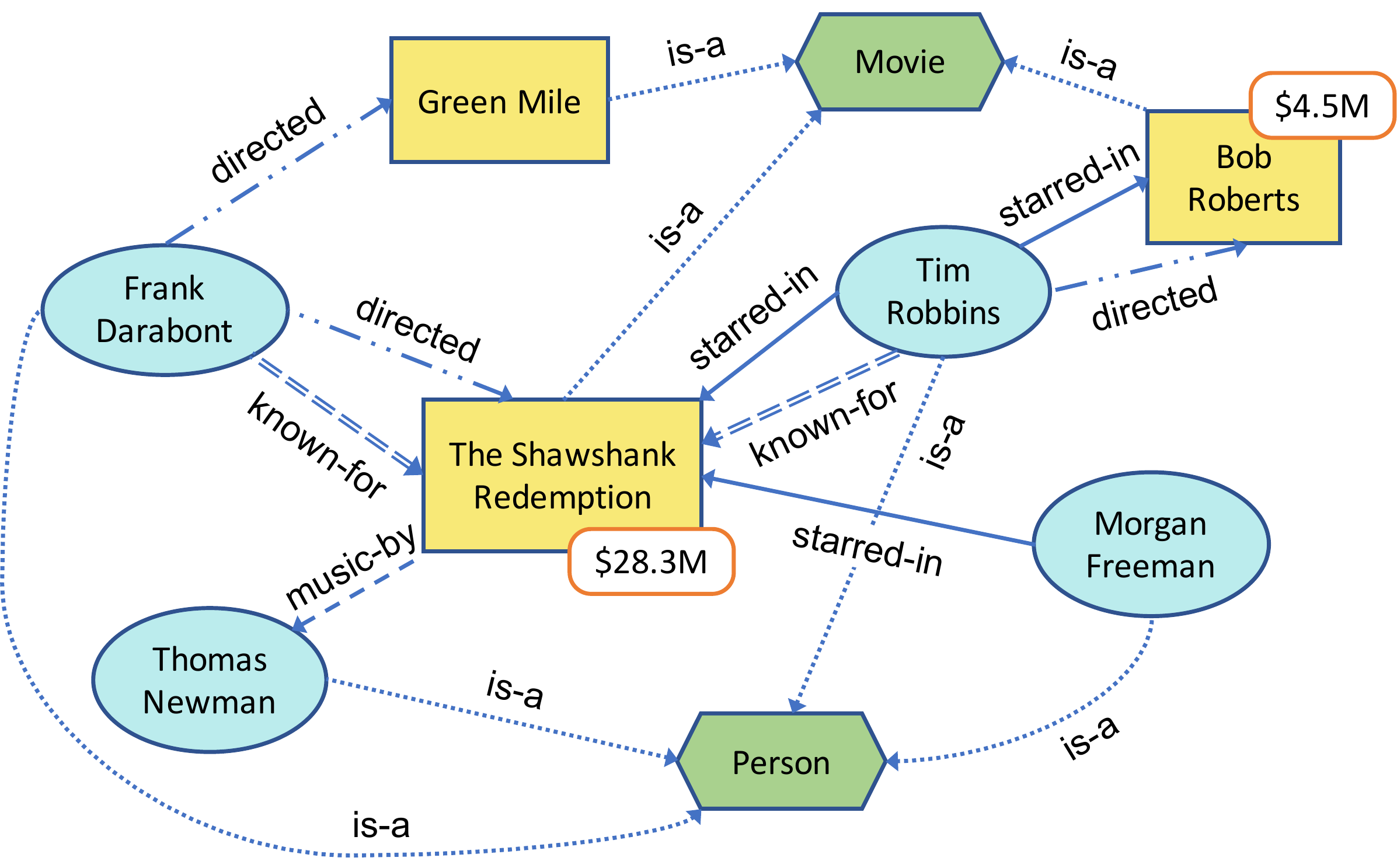}}\caption{An example knowledge graph on movies and related entities.
Different edge types represent different types of relations (e.g., ``directed'' and ``starred-in''),
and different shapes denote different entity types.
Rounded rectangles are importance scores known in advance for some movies.}
\label{fig:movie_kg}
\vspace{-1em}
\end{figure}

Knowledge graphs (KGs) such as Freebase~\cite{DBLP:conf/sigmod/BollackerEPST08}, YAGO~\cite{Suchanek:2007:YCS:1242572.1242667}, and DBpedia~\cite{DBLP:journals/semweb/LehmannIJJKMHMK15}
have proven highly valuable resources for many applications including
question answering~\cite{DBLP:conf/acl/DongWZX15}, 
recommendation~\cite{DBLP:conf/kdd/ZhangYLXM16},
semantic search~\cite{DBLP:conf/icde/BarbosaWY13}, and
knowledge completion~\cite{DBLP:conf/www/WestGMSGL14}.
A KG is a multi-relational graph where nodes correspond to entities, and
edges correspond to relations between the two connected entities.
An edge in a KG represents a fact stored 
in the form of ``$<$subject$>$ $<$predicate$>$ $<$object$>$'',
(e.g., ``$<$Tim Robbins$>$ $<$starred-in$>$ $<$The Shawshank Redemption$>$'').
KGs are different from traditional graphs that have only a single relation;
KGs normally consist of multiple, different relations
that encode heterogeneous information as illustrated by an example movie KG in \Cref{fig:movie_kg}.

Given a KG, estimating the importance of each node is a crucial task 
that enables a number of applications such as recommendation, 
query disambiguation, and resource allocation optimization.
For example, consider a situation where a customer issues a voice query
``Tell me what Genie is'' to a voice assistant backed by a KG.
If the KG contains several entities with such a name, 
the assistant could use their estimated importance to figure out which one to describe.
Furthermore, many KGs are large-scale, often containing millions to billions of 
entities for which the knowledge needs to be enriched or updated to reflect the current state. 
As validating information in KGs requires a lot of resources due to their size and complexity,
node importance can be used to guide the system to allocate limited resources for entities of high importance.

How can we estimate the importance of nodes in a KG?
In this paper, we focus on the setting where we are given importance scores of some nodes in a KG.
An importance score is a value that represents the significance or popularity of a node in the KG.
For example, the number of pageviews of a Wikipedia page can be used as an importance score of the corresponding entity in a KG
since important nodes tend to attract a lot of attention and search traffic.
Then given a KG, how can we predict node importance
by making use of importance scores known for some nodes
along with auxiliary information in KGs such as edge types (predicates)?

\begin{figure}[!t]
\centering
\includegraphics[width=0.95\linewidth]{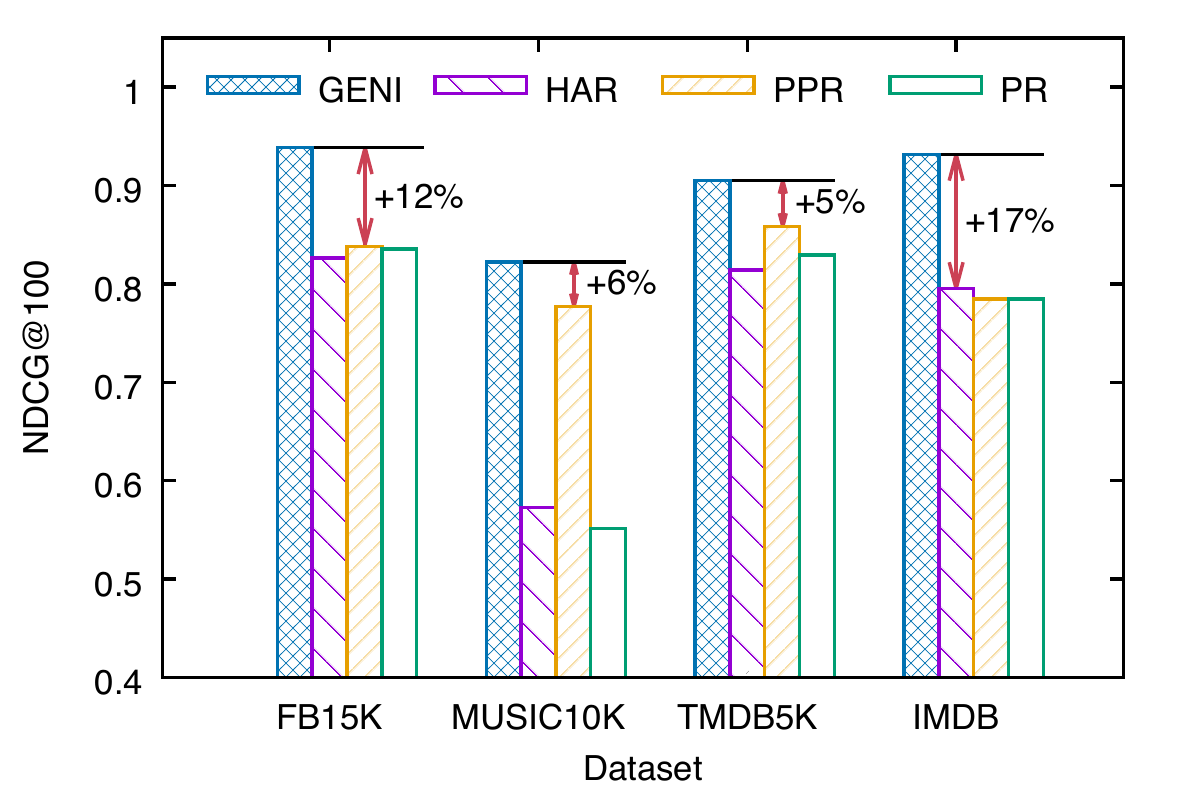}
\caption{Our method \method outperforms existing methods in predicting importance of nodes in real-world KGs.
Higher values are better. See \Cref{sec:exp:realworld} and \Cref{tab:exp:realworld:indomain} for details.}
\label{fig:method_summary}
\vspace{-2em}
\end{figure}

In the past, several approaches have been developed for node importance estimation.
PageRank (\pr)~\cite{page1999pagerank} is an early work on this problem that revolutionized the field of Web search.
However, \pr scores are based only on the graph structure, and unaware of importance scores available for some nodes.
Personalized PageRank (\ppr)~\cite{DBLP:conf/www/Haveliwala02} dealt with this limitation 
by letting users provide their own notion of node importance in a graph.
\ppr, however, does not take edge types into account.
\har~\cite{DBLP:conf/sdm/LiNY12} extends ideas used by \pr and \ppr to distinguish between different predicates in KGs 
while being aware of importance scores and graph topology.
Still, we observe that there is much room for improvement, as evidenced by the performance of existing methods on real-world KGs in \Cref{fig:method_summary}.
So far, existing techniques have approached this problem in a non-trainable framework
that is based on a fixed model structure determined by their prior assumptions on the propagation of node importance, and
involve no learnable parameters that are optimized based on the ground truth.

In this paper, we explore a new family of solutions for the task of predicting node importance in KGs,
namely, regularized supervised machine learning algorithms.
Our goal is to develop a more flexible supervised approach that learns from ground truth, and makes use of additional information in KGs.
Among several supervised algorithms we explore, we focus on graph neural networks (GNNs).
Recently, GNNs have received increasing interests,
and achieved state-of-the-art performance on node and graph classification tasks across data drawn from several domains~\cite{DBLP:journals/corr/KipfW16,DBLP:conf/nips/DefferrardBV16,DBLP:conf/nips/HamiltonYL17,DBLP:conf/kdd/YingHCEHL18,gat2018}.
Designed to learn from graph-structured data, and based on neighborhood aggregation framework,
GNNs have the potential to make further improvements over earlier approaches.
However, existing GNNs have focused on graph representation learning via embedding aggregation, and 
have not been designed to tackle challenges that arise with supervised estimation of node importance in KGs.
Challenges include 
modeling the relationship between the importance of neighboring nodes, 
accurate estimation that generalizes across different types of entities, and
incorporating prior assumptions on node importance that aid model prediction, 
which are not addressed at the same time by existing supervised techniques.

We present \method, a GNN for Estimating Node Importance in KGs.
\method applies an attentive GNN for predicate-aware score aggregation
to capture relations between the importance of nodes and their neighbors. 
\method also allows flexible score adjustment according to node \emph{centrality}, 
which captures connectivity of a node in terms of graph topology.
Our main contributions are as follows.
\begin{compactitem}
\item 
We explore regularized supervised machine learning algorithms for estimating node importance in KGs, 
as opposed to non-trainable solutions where existing approaches belong.
\item We present \method, a GNN-based method designed to address the challenges involved with 
supervised estimation of node importance in KGs.
\item We provide empirical evidence and an analysis of \method using real-world KGs.
\Cref{fig:method_summary} shows that \method outperforms the state of the art by 5\%-17\% percentage points on real KGs.
\end{compactitem}

The rest of this paper is organized as follows.
We present preliminaries in \Cref{sec:prelim}, and describe our method in \Cref{sec:method}.
After providing experimental results on real KGs in \Cref{sec:exp}, 
we review related works in \Cref{sec:related}, and conclude in \Cref{sec:concl}.

\vspace{-0.5em}
\section{Preliminaries}
\label{sec:prelim}
\subsection{Problem Definition} 
\label{sec:prelim:problem}

A \textit{knowledge graph} (KG) is a graph $ G = (V, E = \{ E_1, E_2, \ldots, E_P  \}) $ 
that represents multi-relational data where nodes $ V $ and edges $ E $ correspond to entities and their relationships, respectively;
$ P $ is the number of types of edges (predicates); and
$ E_p $ denotes a set of edges of type $ p \in \{ 1, \ldots, P \} $.
In KGs, there are often many types of predicates (i.e., $ P \gg 1 $) between nodes of possibly different types (e.g., movie, actor, and director nodes), 
whereas in traditional graphs, nodes are connected by just one type of edges (i.e., $ P = 1 $).

An \textit{importance score} $ s \in \mathbb{R}_{\ge 0} $ is a non-negative real number
that represents the significance or popularity of a node.
For example, the total gross of a movie can be used as an importance score for a movie KG,
and the number of pageviews of an entity can be used in a more generic KG such as Freebase~\cite{DBLP:conf/sigmod/BollackerEPST08}.
We assume a single set of importance scores, so the scores can compare with each other to reflect importance.

We now define the node importance estimation problem.
\begin{definition}[Node Importance Estimation]\label{def:problem}
Given a KG $ G = (V, E = \{ E_1, E_2, \ldots, E_P  \}) $ and importance scores $ \{ s \} $ for a subset $ \nodeWithScores \subseteq V $ of nodes,
learn a function $ S: V \rightarrow [0, \infty) $ that 
estimates the importance score of every node in KG.
\end{definition}
\Cref{fig:movie_kg} shows an example KG on movies and related 
entities with importance scores given in advance for some movies.
We approach the importance estimation problem
by developing a supervised framework learning a function 
that maps any node in KG to its score, 
such that the estimation reflects its true importance as closely as possible.

Note that even when importance scores are provided for only one type of nodes (e.g., movies), 
we aim to do estimation for all types of nodes (e.g,. directors, actors, etc.).
\begin{definition}[In-Domain and Out-Of-Domain Estimation]\label{def:estimationtypes}
Given importance scores for some nodes $ \nodeWithScores \subseteq V $ of type $ \mathcal{T} $ (e.g., movies),
predicting the importance of nodes of type $ \mathcal{T} $ is called an \textit{``in-domain''} estimation, and 
importance estimation for those nodes whose type is not $ \mathcal{T} $ is called an \textit{``out-of-domain''} estimation.
\end{definition}
As available importance scores are often limited in terms of numbers and types,
developing a method that generalizes well for both classes of estimation is an important challenge for supervised node importance estimation.

\begin{table}[!t]
\tablefont
\centering
\caption{Comparison of methods for estimating node importance.
\textit{Neighborhood}: Neighborhood awareness. 
\textit{Predicate}: Making use of predicates. 
\textit{Centrality}: Centrality awareness. 
\textit{Input Score}: Utilizing input importance scores. \textit{}
\textit{Flexibility}: Flexible adaptation.}
\centering
\begin{tabular}{ c | c | c | c | c }
\toprule
& \textbf{\method} & \textbf{\har}~\cite{DBLP:conf/sdm/LiNY12} & \textbf{\ppr}~\cite{DBLP:conf/www/Haveliwala02} & \textbf{\pr}~\cite{page1999pagerank} \\
\midrule
\textit{Neighborhood} & $\pmb{\checkmark}$ & $\checkmark$ & $\checkmark$ & $\checkmark$ \\
\textit{Predicate} & $\pmb{\checkmark}$ & $\checkmark$ & & \\
\textit{Centrality} & $\pmb{\checkmark}$ & $\checkmark$ & $\checkmark$ & $ \checkmark $ \\
\textit{Input Score} & $\pmb{\checkmark}$ & $\checkmark$ & $\checkmark$ & \\
\textit{Flexibility} & $\pmb{\checkmark}$ & & & \\
\bottomrule
\end{tabular}

\label{tab:salesmatrix}
\vspace{-2em}
\end{table}

\subsection{Desiderata for Modeling Node Importance in KGs}
\label{sec:prelim:desiderata}

Based on our discussion on prior approaches (\pr, \ppr, and \har), 
we present the desiderata that have guided the development of our method
for tackling node importance estimation problem. 
\Cref{tab:salesmatrix} summarizes \method and existing methods in terms of these desiderata.

\textbf{\textit{Neighborhood Awareness}.}
In a graph, a node is connected to other nodes, except for the special case of isolated nodes.
As neighboring entities interact with each other, and they tend to share common characteristics (network homophily),
neighborhoods should be taken into account when node importance is modeled. 

\textbf{\textit{Making Use of Predicates}.}
KGs consist of multiple types of predicates.
Under the assumption that different predicates could play 
a different role in determining node importance,
models should make predictions using information from predicates.

\textbf{\textit{Centrality Awareness}.}
Without any other information, it is reasonable to assume that 
highly central nodes are more important than less central ones.
Therefore, scores need to be estimated in consideration of node centrality,
capturing connectivity of a node.

\textbf{\textit{Utilizing Input Importance Scores}.}
In addition to graph topology, 
input importance scores provide valuable information to infer relationships between nodes and their importance.
Thus, models should tap into both the graph structure and input scores for more accurate prediction.

\textbf{\textit{Flexible Adaptation}.}
Our assumption regarding node importance such as the one on centrality 
may not conform to the real distribution of input scores over KGs.
Also, we do not limit models to a specific type of input scores.
On the other hand, models can be provided with input scores that possess different characteristics.
It is thus critical that a model can flexibly adapt to the importance that input scores reflect.

\vspace{-1em}
\subsection{Graph Neural Networks}
\label{sec:prelim:gnn}
In this section, we present a generic definition of graph neural networks (GNNs).
GNNs are mainly based on 
neighborhood aggregation architecture~\cite{DBLP:journals/corr/KipfW16,DBLP:conf/nips/HamiltonYL17,DBLP:conf/icml/GilmerSRVD17,DBLP:conf/kdd/YingHCEHL18,gat2018}.
In a GNN with $ L $ layers,
its $ \ell $-th layer ($ \ell = 1, \ldots, L $) receives a feature vector $ \vec{h}^{\ell-1}_i $ for each node $ i $ from the $ (\ell-1) $-th layer (where $ \vec{h}^{0}_i $ is an input node feature $ \vec{z}_i $),
and updates it by aggregating feature vectors from the neighborhood $ \N(i) $ of node $ i $, possibly using a different weight $w^{\ell}_{i,j}$ for neighbor~$ j $.
As updated feature vectors become the input to the $ (\ell + 1) $-th layer,
repeated aggregation procedure through $ L $ layers in principle captures $ L $-th order neighbors in learning a node's representation.
This process of learning representation $ \vec{h}^{\ell}_{i} $ of node $ i $ by $ \ell $-th layer 
is commonly expressed as~\cite{DBLP:conf/nips/HamiltonYL17,DBLP:conf/kdd/YingHCEHL18,DBLP:conf/icml/XuLTSKJ18}:
\begin{align} \label{eq:0}
& \vec{h}^{\ell}_{\N(i)} \leftarrow \textsc{Transform}^{\ell} \left( \textsc{Aggregate} \left( \left\{ \left( \vec{h}^{\ell-1}_j, w^{\ell}_{i,j} \right) ~ \big| ~ j \in \N(i) \right\} \right) \right) \\ \label{eq:1}
& \vec{h}^{\ell}_{i} \leftarrow \textsc{Combine} \left( \vec{h}^{\ell-1}_{i}, ~ \vec{h}^{\ell}_{\N(i)} \right)
\end{align}
where \textsc{Aggregate} is an aggregation function defined by the model (e.g., averaging or max-pooling operation);
\textsc{Transform} is a model-specific function that performs a (non-linear) transformation of node embeddings 
via parameters in $ \ell $-th layer shared by all nodes (e.g., multiplication with a shared weight matrix $ \mat{W^{\ell}} $ followed by some non-linearity $ \sigma(\cdot) $);
\textsc{Combine} is a function that merges the aggregated neighborhood representation with the node's representation (e.g., concatenation).

\eat{
The GNN framework naturally allows us to utilize input importance scores to train a model with flexible adaptation. Its propagation mechanism also allows us to be neighborhood aware. We further enhance the model in three ways.

\begin{compactitem}
\item \textit{Neighborhood Importance Awareness}: GNN normally propagates information between neighbors through node embedding. This is to model the assumption that an entity and its neighbors affect each other, and thus the representation of an entity can be better represented in terms of the representation of its neighbors. In the context of node importance estimation, neighboring importance scores play a major role on the importance of a node, whereas other neighboring features has little effect, if any. We thus directly aggregate importance scores from neighbors (\Cref{sec:method:scoreaggregation}). \item \textit{Making Use of Predicates}:
We design predicate-aware attention mechanism that models how predicates affect the importance of connected entities (\Cref{sec:method:attention}).
\item \textit{Centrality Awareness}:
We apply centrality adjustment to incorporate node centrality into the estimation (\Cref{sec:method:centrality}).
\end{compactitem}

We next describe our GENI model in detail.
}

\vspace{-1.5em}
\section{Method}
\label{sec:method}
Effective estimation of node importance in KGs involves addressing the requirements presented in \Cref{sec:prelim:desiderata}.
As a supervised learning method,
the GNN framework naturally allows us to \textit{utilize input importance scores} to train a model with \textit{flexible adaptation}. 
Its propagation mechanism also allows us to be \textit{neighborhood aware}. In this section, we present \method, which further enhances the model in three ways.
\begin{compactitem}
	\item \textit{Neighborhood Importance Awareness}: GNN normally propagates information between neighbors through node embedding. This is to model the assumption that an entity and its neighbors affect each other, and thus the representation of an entity can be better represented in terms of the representation of its neighbors. In the context of node importance estimation, neighboring importance scores play a major role on the importance of a node, whereas other neighboring features may have little effect, if any. 
	We thus directly aggregate importance scores from neighbors (\Cref{sec:method:scoreaggregation}), and 
	show empirically that it outperforms embedding propagation (\Cref{sec:exp:realworld}).
	\item \textit{Making Use of Predicates}:
	We design predicate-aware attention mechanism that models how predicates affect the importance of connected entities (\Cref{sec:method:attention}).
	\item \textit{Centrality Awareness}:
	We apply centrality adjustment to incorporate node centrality into the estimation (\Cref{sec:method:centrality}).
\end{compactitem}

\begin{figure*}[!t]
\vspace{-1.5em}
\centering
\begin{subfigure}[t]{0.55\textwidth}
	\centering
	\includegraphics[width=.90\linewidth]{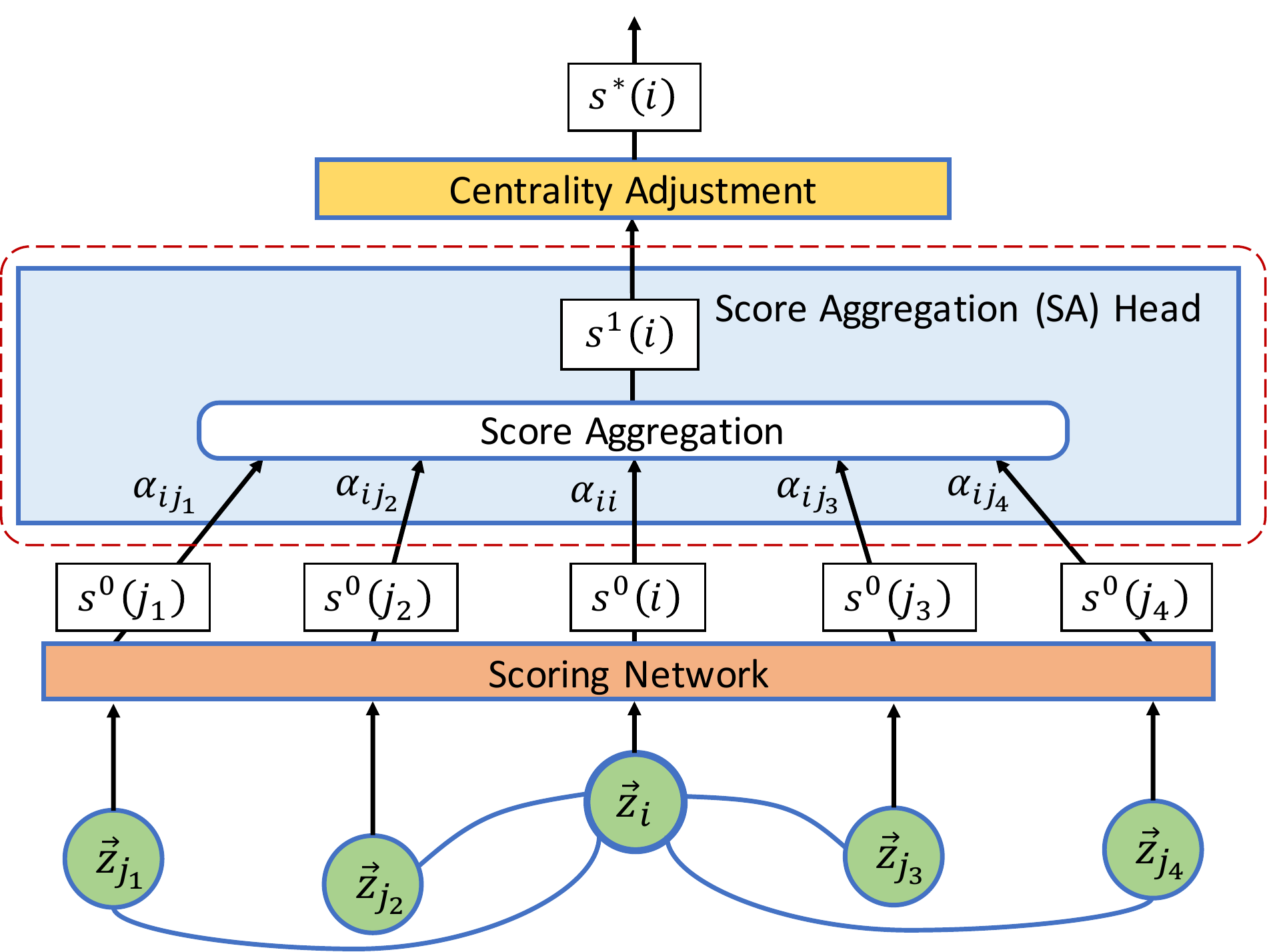}
\caption{}
	\label{fig:model1}
\end{subfigure}
\begin{subfigure}[t]{0.44\textwidth}
	\centering
	\includegraphics[width=.98\linewidth]{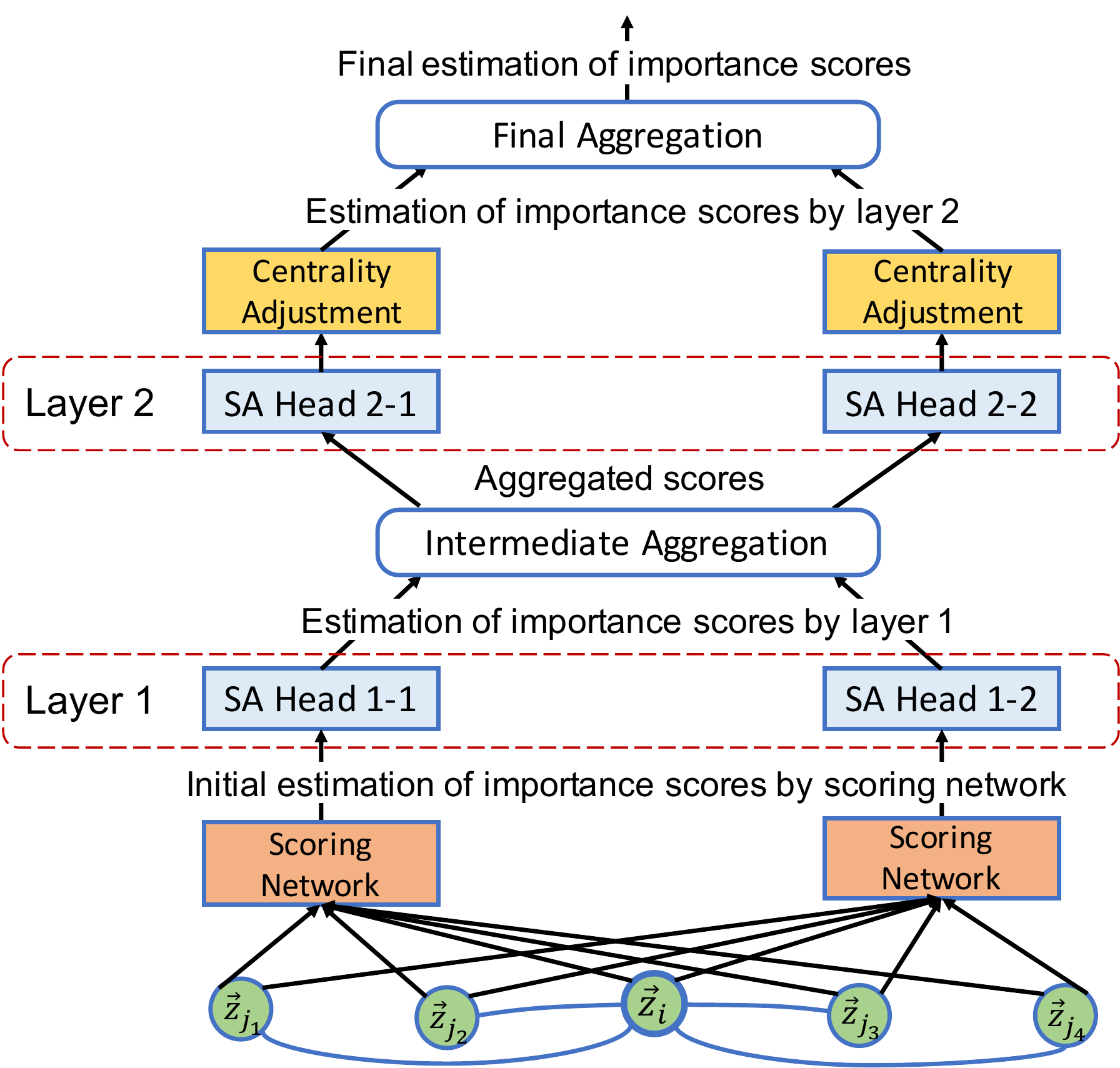}
	\caption{}
	\label{fig:model2}
\end{subfigure}
\vspace{-1.25\baselineskip}
\caption{Description of node importance estimation by \method. 
\subref{fig:model1}: Estimation of the importance of node $ i $ based on the embeddings of node $ i $ and 
its neighbors $ j_1, \ldots, j_4 $ (connected by blue edges).
The final estimation $ s^{*}(i) $ is produced via three components of \method shown in colored boxes, which are described in \Cref{sec:method:scoreaggregation,sec:method:attention,sec:method:centrality}.
\subref{fig:model2}: An illustration of the proposed model that consists of two layers, each of which contains two score aggregation heads.
Note that the model can consist of different numbers of layers, and each layer can also have different numbers of score aggregation heads.
A discussion on the extension of the basic model in \subref{fig:model1} to 
a more comprehensive architecture in \subref{fig:model2} is given in \Cref{sec:method:architecture}.}
\label{fig:model}
\vspace{-1em}
\end{figure*}

An overview of \method is provided in \Cref{fig:model}.
In \Cref{sec:method:scoreaggregation,sec:method:attention,sec:method:centrality},
we describe the three main enhancements using the basic building blocks of \method shown in \Cref{fig:model1}.
Then we discuss an extension to a general architecture in \Cref{sec:method:architecture}.
\Cref{tab:symbols} provides the definition of symbols used in this paper.

\begin{table}[!t]
\vspace{-1mm}
\small
\caption{Table of symbols.}
\centering
\makebox[0.4\textwidth][c]{
\setlength{\tabcolsep}{0.75mm}
\begin{tabular}{ c | l}
\toprule
\textbf{Symbol} & \textbf{\makecell{Definition}} \\
\midrule
$ \nodeWithScores $ & set of nodes with known importance scores \\
$ \vec{z}_i $ & real-valued feature vector of node $ i $ \\
$ \N(i) $ & neighbors of node $ i $ \\
$ L $ & total number of score aggregation (SA) layers \\
$ \ell $ & index for an SA layer \\
$ H^{\ell} $ & number of SA heads in $ \ell $-th layer \\
$ p_{ij}^{m} $ & predicate of $ m $-th edge between nodes $ i $ and $ j $ \\
$ \phi(e) $ & learnable embedding of predicate $ e $ \\
$ \sigma_a, \sigma_s $ & non-linearities for attention computation and score estimation \\
$ s_h^{\ell}(i) $ & \makecell[l]{estimated score of node $ i $ by $ h $-th SA head in $ \ell $-th layer} \\
$ s^{*}(i) $ & centrality-adjusted score estimation of node $ i $\\
$ || $ & concatenation operator \\
$ d(i) $ & in-degree of node $ i $ \\
$ c(i) $ & centrality score of node $ i $ \\
$ c^*_h(i) $ & centrality score of node $ i $ scaled and shifted by $ h $-th SA head \\
$ \gamma_h, \beta_h $ & learnable scale and shift parameters used by $ h $-th SA head \\
$\vec{a}_{h,\ell}$ & learnable parameter vector to compute $\alpha^{h,\ell}_{ij}$ by $ h $-th SA head in $ \ell $-th layer \\
$ \alpha^{h,\ell}_{ij} $ & \makecell[l]{node $ i $'s attention on node $ j $ computed with $ h $-th SA head in $ \ell $-th layer} \\
$ g(i) $ & known importance score of node $ i $\\
\bottomrule
\end{tabular}
}	
\label{tab:symbols}
\vspace{-2.5em}
\end{table}

\subsection{Score Aggregation}
\label{sec:method:scoreaggregation}

\eat{
Most existing GNNs are structured around a common neighborhood aggregation framework
where they aggregate embeddings of neighboring nodes for graph representation learning.
This is to model the assumption that an entity and its neighbors affect each other, and thus 
the representation of an entity can be better represented in terms of the representation of its neighbors.
However, in the context of node importance estimation,
this architecture performs the final estimation based on aggregated node embeddings, 
and does not explicitly consider how the importance of an entity is affected by that of neighboring entities.
}

To directly model the relationship between the importance of neighboring nodes, 
we propose a score aggregation framework, rather than embedding aggregation.
Specifically, in~\Cref{eq:0,eq:1}, we replace the hidden embedding $\vec{h}^{\ell-1}_j$ of node $ j $ 
with its score estimation $s^{\ell-1}(j)$ and combine them as follows:
\begin{align}\label{eq:nbraggr}
s^{\ell}(i) = \sum_{j \in \N(i) \cup \{ i \}} \alpha^\ell_{ij} ~ s^{\ell-1}(j)
\end{align}
where $ \N(i) $ denotes the neighbors of node $ i $, which will be a set of the first-order neighbors of node $ i $ in our experiments.
Here, $ \alpha^\ell_{ij} $ is a learnable weight between nodes $ i $ and $ j $ for the $\ell$-th layer ($\ell = 1, \ldots, L$). 
We train it via a shared attention mechanism 
which is computed by a pre-defined model with shared parameters and predicate embeddings, as we explain soon. 
In other words, \method computes the aggregated score $ s^{\ell}(i) $ 
by performing a weighted aggregation of intermediate scores from node $ i $ and its neighbors.
Note that \method does not apply $ \textsc{Transform}^{\ell} $ function after aggregation as in \Cref{eq:0},
since \method aggregates scores.
Propagating scores instead of node embeddings has the additional benefit of reducing the number of model parameters.

\eat{
More specifically, we suppose that the importance of an entity in a KG is
determined by its own state and its relationship with the importance of nearby entities.
To model this interaction between neighboring nodes, 
we distinguish between two types of scores:
an \textit{``intermediate score''} that represents an importance estimation 
based only on the node's current state, 
and an \textit{``aggregated score''} that is obtained
after considering information of neighboring nodes as well.
With intermediate scores $ s'(i) $ computed for each node $ i $,
\method computes the aggregated score $ s(i) $ 
by performing a weighted aggregation of intermediate scores from node $ i $ and its neighbors:
\begin{align}\label{eq:nbraggr}
s(i) = \sum_{j \in \N(i) \cup \{ i \}} \alpha_{ij} ~ s'(j)
\end{align}
where $ \N(i) $ denotes the neighbors of node $ i $, and
$ \alpha_{ij} $ is a learnable weight between nodes $ i $ and $ j $.
In our experiments, we define $ \N(i) $ to be a set of nodes directly connected to node $ i $.
}

To compute the initial estimation $s^0(i)$, \method uses input node features.
In the simplest case, they can be one-hot vectors that represent each node.
More generally, they are real-valued vectors representing the nodes, 
which are extracted manually based on domain knowledge, or generated with methods for learning node embeddings.
Let $\vec{z}_i$ be the input feature vector of node $ i $. 
Then \method computes the initial score of $i$ as \begin{align}
s^0(i) = \textsc{ScoringNetwork}(\vec{z}_i)
\end{align}
where \textsc{ScoringNetwork} can be any neural network that takes in a node 
feature vector and returns an estimation of its importance.
We used a simple fully-connected neural network for our experiments.

\subsection{Predicate-Aware Attention Mechanism}
\label{sec:method:attention}

Inspired by recent work that showcased successful application of attention
mechanism, we employ a predicate-aware attention mechanism
that attends over the neighbor's intermediate scores.

Our attention considers two factors. First, we consider the predicate between the nodes because different predicates can play different roles for score propagation. For example, even though a movie may be released in a popular (i.e., important) country, the movie itself may not be popular; on the other hand, a movie directed by a famous (i.e., important) director is more likely to be popular. Second, we consider the neighboring score itself in deciding the attention. 
A director who directed a few famous (i.e., important) movies is likely to be important; the fact that he also directed some not-so-famous movies in his life is less likely to make him unimportant.  

\eat{
Weighted aggregation reflects the fact that influence from neighboring
nodes are not all the same, particularly in KGs.
This is where we find the type of relation (predicate) between two entities beneficial.
As an example, consider a movie KG where we have nodes 
corresponding to countries, and movie nodes are connected to the country
nodes in which they were produced.
Although country nodes are extremely central,
which country a movie was produced in should not play a major role
in predicting movie popularity.

Inspired by recent work that showcased successful application of attention
mechanism, we employ a predicate-aware attention mechanism
that attends over the neighbor's intermediate scores
with consideration of predicates.
}

\method incorporates predicates into attention computation
by using shared predicate embeddings; i.e., 
each predicate is represented by a feature vector of predefined length, and
this representation is shared by nodes across all layers. 
Further, predicate embeddings are learned so as to
maximize the predictive performance of the model in a flexible fashion.
Note that in KGs, there could be multiple edges of 
different types between two nodes (e.g., see \Cref{fig:movie_kg}).
We use $ p_{ij}^{m} $ to denote the predicate of $ m $-th edge 
between nodes $ i $ and $ j $, and $ \phi(\cdot) $ to denote a
mapping from a predicate to its embedding.

In \method, we use a simple, shared self-attention mechanism,
which is a single layer feedforward neural network 
parameterized by the weight vector $\vec{a}$.
Relation between the intermediate scores of two nodes $ i $ and $ j $,
and the role an in-between predicate plays are captured 
by the attentional layer that takes in the concatenation of
all relevant information.
Outputs from the attentional layer are first transformed by 
non-linearity $ \sigma(\cdot) $, and then normalized via the softmax function.
Formally, \method computes the attention $ \alpha_{ij}^{\ell} $ of node $ i $ on node $ j $ for $ \ell $-th layer as:
\begin{align}
\alpha^\ell_{ij} = \frac{
\exp \left( \sigma_a \left( \sum_{m} \vec{a}_{\ell}^\top [ s^\ell(i) || \phi(p_{ij}^{m}) || s^\ell(j) ] \right) \right) 
}{
\sum_{k \in \N(i) \cup \{ i \}} \exp \left( \sigma_a \left( \sum_{m} \vec{a}_{\ell}^\top [ s^\ell(i) || \phi(p_{ik}^{m}) || s^\ell(k) ] \right) \right)
}
\label{eq:simple:attention}
\end{align}
where $ \sigma_a $ is a non-linearity, $ \vec{a}_{\ell} $ is a weight vector for $ \ell $-th layer, 
and $ || $ is a concatenation operator.

\subsection{Centrality Adjustment}
\label{sec:method:centrality}

Existing methods such as PR, PPR, and HAR make a common assumption that 
the importance of a node positively correlates with its centrality in the graph.
In the context of KGs, it is also natural to assume that more central nodes would be more important than less central ones,
unless the given importance scores present contradictory evidence.
Making use of this prior knowledge becomes especially beneficial 
in cases where we are given a small number of importance scores compared to the total number of entities, and in cases where
the importance scores are given for entities of a specific type out of the many types in KG.

Given that the in-degree $ d(i) $ of node $ i $ is a common proxy for its centrality and popularity,
we define the initial centrality $ c(i) $ of node $ i $ to be
\begin{align}
c(i) = \log(d(i)+\epsilon)
\end{align}
where 
$ \epsilon $ is a small positive constant.

While node centrality provides useful information on the importance of a node, strictly adhering to the node centrality could have a detrimental effect on model prediction.
We need flexibility to account for the possible discrepancy between the node's centrality in a given KG and 
the provided input importance score of the node.
To this end, we use a scaled and shifted centrality $ c^{*}(i) $ as our notion of node centrality:
\begin{align}\label{eq:flexible_ca}
c^{*}(i) = \gamma \cdot c(i) + \beta
\end{align}
where $ \gamma $ and $ \beta $ are learnable parameters for scaling and shifting. 
As we show in Section~\ref{sec:exp:analysis}, this flexibility allows better performance when in-degree is not the best proxy of centrality.

To compute the final score, we apply centrality adjustment to the score estimation $ s^L(i) $ from the last layer, and apply a non-linearity $ \sigma_s $
as follows:
\begin{align}\label{eq:final_estimation}
s^*(i) = \sigma_s \left( c^*(i) \cdot s^L(i) \right)
\end{align}

\subsection{Model Architecture}
\label{sec:method:architecture}

The simple architecture depicted in \Cref{fig:model1} consists of a scoring network and a single score aggregation (SA) layer (i.e., $L=1$), 
followed by a centrality adjustment component. \Cref{fig:model2}
extends it to a more general architecture in two ways. 
First, we extend the framework to contain multiple SA layers; that is, $L>1$. 
As a single SA layer aggregates the scores of direct neighbors, stacking multiple SA layers enables aggregating scores from a larger neighborhood. 
Second, we design each SA layer to contain a variable number of SA heads, 
which perform score aggregation and attention computation independently of each other.
Empirically, we find using multiple SA heads to be helpful for the model performance and the stability of optimization procedure (Section~\ref{sec:exp:analysis}).

Let $ h $ be an index of an SA head, and $ H^{\ell} $ be the number of SA heads in $ \ell $-th layer.
We define $ s'^{\ell-1}_h(i) $ to be node $ i $'s score that is estimated by $ (\ell-1) $-th layer, and
fed into $ h $-th SA head in $ \ell $-th (i.e., the next) layer, 
which in turn produces an aggregation $ s_{h}^{\ell}(i) $ of these scores:
\begin{align}
s^{\ell}_h(i) = \sum_{j \in \N(i) \cup \{ i \}} \alpha_{ij}^{h,\ell} ~ s'^{\ell-1}_h(j)
\end{align}
where $ \alpha_{ij}^{h,\ell} $ is the attention coefficient between nodes $ i $ and $ j $ computed by SA head $ h $ in layer $ \ell $.

In the first SA layer, each SA head $ h $ receives input scores from a separate scoring network $ \textsc{ScoringNetwork}_h $, 
which provides the initial estimation $ s_h^0(i) $ of node importance.
For the following layers, output from the previous SA layer becomes the input estimation.
Since in $ \ell $-th $ (\ell \ge 1) $ SA layer, $ H^{\ell} $ SA heads independently produce $ H^{\ell} $ score estimations in total,
we perform an aggregation of these scores by averaging, which is provided to the next layer. That is,
\begin{align}
s'^{\ell}_{h}(i) = \begin{cases}
\textsc{ScoringNetwork}_h(\vec{z}_i) & \text{if } \ell = 0 \\
\textsc{Average}\left( \left\{ s^{\ell}_{h}(i) ~ \big\rvert ~ h = 1,\ldots,H^{\ell} \right\} \right) & \text{if } \ell \ge 1 \\
\end{cases}
\end{align}

Multiple SA heads in $ \ell $-th layer compute attention between neighboring nodes in the same way as in \Cref{eq:simple:attention},
yet independently of each other using its own parameters $ \vec{a}_{h,\ell} $:
\begin{align}
\alpha_{ij}^{h,\ell} = \frac{
	\exp \left( \sigma_a \left( \sum_{m} \vec{a}_{h,\ell}^\top [ s'^{\ell-1}_{h}(i) || \phi(p_{ij}^{m}) || s'^{\ell-1}_{h}(j) ] \right) \right) 
}{
	\sum_{k \in \N(i) \cup \{ i \}} \exp \left( \sigma_a \left( \sum_{m} \vec{a}_{h,\ell}^\top [ s'^{\ell-1}_{h}(i) || \phi(p_{ik}^{m}) || s'^{\ell-1}_{h}(k) ] \right) \right)
}
\end{align}

Centrality adjustment is applied to the output from the final SA layer.
In order to enable independent scaling and shifting by each SA head,
separate parameters $ \gamma_h $ and $ \beta_h $ are used for each head $ h $.
Then centrality adjustment by $ h $-th SA head in the final layer is:
\begin{align}
c^*_h(i) = \gamma_h \cdot c(i) + \beta_h
\end{align}

With $ H^L $ SA heads in the final $ L $-th layer, 
we perform additional aggregation of centrality-adjusted scores by averaging, and apply a non-linearity $ \sigma_s $,
obtaining the final estimation $ s^*(i) $:
\begin{align}
s^*(i) = \sigma_s \left( \textsc{Average}\left( \left\{ c^*_h(i) \cdot s^L_{h}(i) ~ \big\rvert ~ h = 1,\ldots,H^L \right\} \right) \right)
\end{align}

\subsection{Model Training}
\label{sec:method:modeltraining}

In order to predict node importance with input importance scores known for a subset of nodes $ \nodeWithScores \subseteq V $,
we train \method using mean squared error between the given importance score $ g(i) $ and the model estimation $ s^{*}(i) $ for node $ i \in \nodeWithScores$; thus, the loss function is
\begin{align}
\frac{1}{|\nodeWithScores|} \sum_{i \in \nodeWithScores} \left( s^{*}(i) - g(i) \right)^2
\end{align}
Note that \textsc{ScoringNetwork} is trained jointly with the rest of \method.
To avoid overfitting, we apply weight decay with an early stopping criterion based on the model performance on validation entities.

\section{Experiments}
\label{sec:exp}
In this section, we aim to answer the following questions.
\begin{compactitem}
\item How do \method and baselines perform on real-world KGs with different characteristics?
In particular, how well do methods perform in- and out-of-domain estimation~(\Cref{def:estimationtypes})?
\item How do the components of \method, such as centrality adjustment, and different parameter values affect its estimation?
\end{compactitem}
We describe datasets, baselines, and evaluation plans
in~\Cref{sec:exp:datasets,sec:exp:baselines,sec:exp:evaluation}, and 
answer the above questions in \Cref{sec:exp:realworld,sec:exp:analysis}.

\subsection{Datasets}
\label{sec:exp:datasets}

\begin{table*}[!htbp]
\vspace{-1mm}
\tablefont
\centering
\setlength{\tabcolsep}{1.5mm}
\caption{Real-world KGs. See \Cref{sec:exp:datasets} and \Cref{appendix:datasets} for details. SCC: Strongly connected component. OOD: Out-of-domain.}
\centering
\begin{tabular}{ c | r | r | r | r | c | r | c }
	\toprule
	\textbf{Name} & \textbf{\makecell{\# Nodes}} & \textbf{\makecell{\# Edges}} & \textbf{\makecell{\# Predicates}} & \textbf{\makecell{\# SCCs.}} & \textbf{\makecell{Input Score Type}} & \textbf{\makecell{\# Nodes w/ Scores}} & \textbf{\makecell{Data for OOD Evaluation}} \\
	\midrule
	\fb & 14,951 & 592,213 & 1,345 & 9 & \# Pageviews & 14,108 (94\%) & N/A \\
	\music & 24,830 & 71,846 & 10 & 130 & Song hotttnesss & 4,214 (17\%) & Artist hotttnesss \\
	\tmdb & 123,906 & 532,058 & 22 & 15 & Movie popularity & 4,803 (4\%) & Director ranking \\
	\imdb & 1,567,045 & 14,067,776 & 28 & 1 & \# Votes for movies & 215,769 (14\%) & Director ranking \\
	\bottomrule
\end{tabular}	
\label{tab:datasets}
\vspace{-1em}
\end{table*}

In our experiments, we use four real-world KGs with different characteristics.
Here we introduce these KGs along with the importance scores used for in- and out-of-domain (OOD) evaluations~(see \Cref{def:estimationtypes}).
Summaries of the datasets (such as the number of nodes, edges, and predicates) are given in \Cref{tab:datasets}.
More details such as data sources and how they are constructed can be found in \Cref{appendix:datasets}.

\textbf{\fb} is a subset of Freebase,
which is a large collaborative knowledge base containing general facts, and has been widely used 
for research and practical applications~\cite{DBLP:conf/nips/BordesUGWY13,DBLP:conf/sigmod/BollackerEPST08}.
\fb has a much larger number of predicates and a higher density than other KGs we evaluated.
For each entity, we use the number of pageviews for the corresponding Wikipedia page
as its score. Note that we do not perform OOD evaluation for \fb
since importance scores for \fb apply to all types of entities.

\textbf{\music} is a music KG sampled from the Million Song Dataset\footnote{\url{https://labrosa.ee.columbia.edu/millionsong/}},
which includes information about songs such as the primary artist and the album the song belongs to.
The dataset provides two types of popularity scores called 
``song hotttnesss'' and ``artist hotttnesss'' computed by the Echo Nest platform
by considering data from many sources such as mentions on the web, play counts, etc\footnote{\url{https://musicmachinery.com/tag/hotttnesss/}}.
We use ``song hotttnesss'' as input importance scores, and ``artist hotttnesss'' 
for OOD performance evaluation.

\textbf{\tmdb} is a movie KG derived from the TMDb 5000 movie dataset\footnote{\url{https://www.kaggle.com/tmdb/tmdb-movie-metadata}}.
It contains movies and related entities such as movie genres, companies, countries, crews, and casts.
We use the ``popularity'' information for movies as importance scores, which is provided by the original dataset.
For OOD evaluation, we use a ranking of top-200 highest grossing directors\footnote{\url{https://www.the-numbers.com/box-office-star-records/worldwide/lifetime-specific-technical-role/director}}.
Worldwide box office grosses given in the ranking are used as importance scores for directors.

\textbf{\imdb} is a movie KG created from the public IMDb dataset, which includes
information such as movies, genres, directors, casts, and crews.
\imdb is the largest KG among those we evaluate, with $ 12.6\times $ as many nodes as \tmdb.
IMDb dataset provides the number of votes a movie received, which we use as importance scores.
For OOD evaluation, we use the same director ranking used for \tmdb.

\subsection{Baselines}
\label{sec:exp:baselines}

Methods for node importance estimation in KGs can be classified into two families of algorithms.

\textbf{Non-Trainable Approaches.} Previously developed methods mostly belong to this category. We evaluate the following methods:
\begin{compactitem}
	\item PageRank (\pr)~\cite{page1999pagerank}
	\item Personalized PageRank (\ppr)~\cite{DBLP:conf/www/Haveliwala02}
	\item \har~\cite{DBLP:conf/sdm/LiNY12}
\end{compactitem}

\textbf{Supervised Approaches.} We explore the performance of representative supervised algorithms on node importance estimation:
\begin{compactitem}
	\item Linear regression (\linreg): an ordinary least squares algorithm.
	\item Random forests (\randforest): a random forest regression model.
	\item Neural networks (\neuralnet): a fully-connected neural network. \item Graph attention networks (\gat)~\cite{gat2018}:
	This is a GNN model reviewed in~\Cref{sec:prelim:gnn}.
	We add a final layer that takes the node embedding and outputs the importance score of a node.
	
\end{compactitem}
All these methods and \method use the same data (node features and input importance scores).
In our experiments, node features are generated using node2vec~\cite{DBLP:conf/kdd/GroverL16}.
Depending on the type of KGs, other types of node features, such as bag-of-words representation, can also be used.
Note that the graph structure is explicitly used only by \gat,
although other supervised baselines make an implicit use of it when node features encode graph structural information.

We will denote each method by the name in parentheses.
Experimental settings for baselines and \method are provided in \Cref{appendix:settings}.

\subsection{Performance Evaluation}
\label{sec:exp:evaluation}

We evaluate methods based on their in- and out-of-domain (OOD) performance.
We performed 5-fold cross validation, and report the average and standard deviation of 
the following metrics on ranking quality and correlation: 
normalized discounted cumulative gain and Spearman correlation coefficient.
Higher values are better for all metrics. We now provide their formal definitions.

\textbf{Normalized discounted cumulative gain (NDCG)} is a measure of ranking quality.
Given a list of nodes ranked by predicted scores, and their graded relevance values 
(which are non-negative, real-valued ground truth scores in our setting), 
discounted cumulative gain at position $ k $ ($ DCG@k $) is defined as:
\begin{align}
DCG@k = \sum_{i=1}^{k} \frac{r_i}{\log_2(i + 1)}
\end{align}
where $ r_i $ denotes the graded relevance of the node at position $ i $.
Note that due to the logarithmic reduction factor, the gain $ r_i $ of each node is penalized at lower ranks.
Consider an ideal DCG at rank position $ k $ ($ IDCG@k $) which is obtained by an ideal ordering of nodes based on their relevance scores.
Normalized DCG at position $ k $ ($ NDCG@k $) is then computed as: \begin{align}
NDCG@k = \frac{DCG@k}{IDCG@k}
\end{align}
Our motivation for using $NDCG@k$ is to test the quality of ranking for the top $ k $ entities.

\textbf{Spearman correlation coefficient (\spearman)} measures the rank correlation between the ground truth scores $ \vec{g} $ and predicted scores $ \vec{s} $;
that is, the strength and direction of the monotonic relationship between the rank values of $ \vec{g} $ and $ \vec{s} $.
Converting $ \vec{g} $ and $ \vec{s} $ into ranks $ \vec{g}_r $ and $ \vec{s}_r $, respectively,
Spearman correlation coefficient is computed as:
\begin{align}
Spearman = \frac{\sum_{i}^{} ({g_r}_i - \overline{g}_r)({s_r}_i - \overline{s}_r)}{ \sqrt{\sum_{i} ({g_r}_i - \overline{g}_r)^2} \sqrt{\sum_{i} ({s_r}_i - \overline{s}_r)^2} }
\end{align}
where $ \overline{g}_r $ and $ \overline{s}_r $ are the mean of $ \vec{g}_r $ and $ \vec{s}_r $.

\eat{
\textbf{Average precision} measures the quality of ranking.
For a list of nodes sorted by their predicted scores, precision at $ k $ ($ P@k $) is the proportion of relevant nodes among the top-$ k $ nodes.
Assume that there are $ R $ relevant nodes in the list. 
Denoting the rank position of each relevant node by $ k_i ~ (i=1,\ldots,R) $,
average precision ($ AP $) is defined as:
\begin{align}
AP = \frac{1}{R} \sum_{i=1}^{R} P@k_i
\end{align}
}

For in-domain evaluation, we use \ndcg and \spearman as they complement each other:
NDCG@100 looks at the top-100 predictions, and \spearman considers the ranking of all entities with known scores.
For NDCG, we also tried different cut-off thresholds and observed similar results.
Note that we often have a small volume of data for OOD evaluation.
For example, for \tmdb and \imdb, we used a ranking of 200 directors with known scores,
while \tmdb and \imdb have 2,578 and 287,739 directors, respectively.
Thus \spearman is not suitable for OOD evaluation 
as it considers only those small number of entities in the ranking, and ignores all others, even if they are predicted to be highly important;
thus, for OOD evaluation, we report \ndcg and \ndcgt.

Additionally, we report regression performance in \Cref{appendix:regression_performance}.

\subsection{Importance Estimation on Real-World Data}
\label{sec:exp:realworld}

\begin{table*}[!htbp]
\tablefont
\centering
\setlength{\tabcolsep}{1.0mm}
\caption{
In-domain prediction results on real-world datasets. \method consistently outperforms all baselines.
Numbers after $ \pm $~symbol are standard deviation from 5-fold cross validation.
Best results are in bold, and second best results are underlined.
}
\makebox[\textwidth][c]{
\begin{tabular}{ c | r | r | r | r | r | r | r | r}
\toprule
\multirow{2}{*}{\textbf{Method}} & \multicolumn{2}{c|}{\textbf{\fb}} & \multicolumn{2}{c|}{\textbf{\music}} & \multicolumn{2}{c|}{\textbf{\tmdb}} & \multicolumn{2}{c}{\textbf{\imdb}} \\
& \makecell{\ndcg} & \makecell{\spearman} & \makecell{\ndcg} & \makecell{\spearman} & \makecell{\ndcg} & \makecell{\spearman} & \makecell{\ndcg} & \makecell{\spearman} \\
\midrule
\pr & $ 0.8354 \pm 0.016 $ & $ 0.3515 \pm 0.015 $ & $ 0.5510 \pm 0.021 $ & $ -0.0926 \pm 0.034 $ & $ 0.8293 \pm 0.026 $ & $ 0.5901 \pm 0.011 $ & $ 0.7847 \pm 0.048 $ & $ 0.0881 \pm 0.004 $ \\
\ppr & $ 0.8377 \pm 0.015 $ & $ 0.3667 \pm 0.015 $ & $ 0.7768 \pm 0.009 $ & $ 0.3524 \pm 0.046 $ & $ 0.8584 \pm 0.013 $ & $ \underline{0.7385 \pm 0.010} $ & $ 0.7847 \pm 0.048 $ & $ 0.0881 \pm 0.004 $ \\
\har & $ 0.8261 \pm 0.005 $ & $ 0.2020 \pm 0.012 $ & $ 0.5727 \pm 0.017 $ & $ 0.0324 \pm 0.044 $ & $ 0.8141 \pm 0.021 $ & $ 0.4976 \pm 0.014 $ & $ 0.7952 \pm 0.036 $ & $ 0.1318 \pm 0.005 $ \\ \hline
\linreg & $ 0.8750 \pm 0.005 $ & $ 0.4626 \pm 0.019 $ & $ 0.7301 \pm 0.023 $ & $ 0.3069 \pm 0.032 $ & $ 0.8743 \pm 0.015 $ & $ 0.6881 \pm 0.013 $ & $ 0.7365 \pm 0.009 $ & $ 0.5013 \pm 0.002 $ \\
\randforest & $ 0.8734 \pm 0.005 $ & $ 0.5122 \pm 0.019 $ & $ \underline{0.8129 \pm 0.012} $ & $ \underline{0.4577 \pm 0.012} $ & $ 0.8503 \pm 0.016 $ & $ 0.5959 \pm 0.022 $ & $ 0.7651 \pm 0.010 $ & $ 0.4753 \pm 0.005 $ \\
\neuralnet & $ 0.9003 \pm 0.005 $ & $ 0.6031 \pm 0.012 $ & $ 0.8015 \pm 0.017 $ & $ 0.4491 \pm 0.027 $ & $ 0.8715 \pm 0.006 $ & $ 0.7009 \pm 0.009 $ & $ 0.8850 \pm 0.016 $ & $ 0.5120 \pm 0.008 $ \\
\gat & $ \underline{0.9205 \pm 0.009} $ & $ \underline{0.7054 \pm 0.013} $ & $ 0.7666 \pm 0.016 $ & $ 0.4276 \pm 0.023 $ & $ \underline{0.8865 \pm 0.011} $ & $ 0.7180 \pm 0.010 $ & $ \underline{0.9110 \pm 0.011} $ & $ \underline{0.7060 \pm 0.007} $ \\
\midrule
\textbf{\method} & $ \textbf{0.9385} \pm \textbf{0.004}$ & $\textbf{0.7772} \pm \textbf{0.006}$ & $ \textbf{0.8224} \pm \textbf{0.018} $ & $ \textbf{0.4783} \pm \textbf{0.009} $ & $ \textbf{0.9051} \pm \textbf{0.005} $ & $ \textbf{0.7796} \pm \textbf{0.009} $ & $ \textbf{0.9318} \pm \textbf{0.005} $ & $ \textbf{0.7387} \pm \textbf{0.002} $ \\
\bottomrule
\end{tabular}
}
\label{tab:exp:realworld:indomain}
\end{table*}

We evaluate \method and baselines in terms of in- and out-of-domain (OOD) predictive performance.

\subsubsection{In-Domain Prediction}

\Cref{tab:exp:realworld:indomain} summarizes in-domain prediction performance.
\method outperforms all baselines on four datasets in terms of both \ndcg and \spearman.
It is noteworthy that supervised approaches generally perform better in-domain prediction than non-trainable ones, 
especially on \fb and \imdb, which are more complex and larger than the other two.
It demonstrates the applicability of supervised models to our problem.
On all KGs except \music, \gat outperforms other supervised baselines,
which use the same node features but do not explicitly take the graph network structure into account.
This shows the benefit of directly utilizing network connectivity.
By modeling the relation between scores of neighboring entities, 
\method achieves further performance improvement over \gat.
Among non-trainable baselines, \har often performs worse than \pr and \ppr,
which suggests that considering predicates could hurt performance if predicate weight adjustment is not done properly.

\subsubsection{Out-Of-Domain Prediction}

\Cref{tab:exp:realworld:outofdomain} summarizes OOD prediction results.
\method achieves the best results for all KGs in terms of both \ndcg and \ndcgt.
In contrast to in-domain prediction where supervised baselines generally outperform non-trainable ones,
we observe that non-trainable methods achieve higher OOD results than supervised baselines on \music and \tmdb.
In these KGs, only about 4,000 entities have known scores.
Given scarce ground truth, non-trainable baselines could perform better by relying on a prior assumption on the propagation of node importance.
Further, note that the difference between non-trainable and supervised baselines is more drastic on \tmdb 
where the proportion of nodes with scores is the smallest ($ 4\% $).
On the other hand, on \imdb, which is our largest KG with the greatest number of ground truth, supervised baselines mostly outperform non-trainable methods.
In particular, none of the top-100 directors in \imdb predicted by \pr and \ppr belong to the ground truth director ranking.
With 14\% of nodes in \imdb associated with known scores, supervised methods learn to generalize better for OOD prediction.
Although neighborhood aware, \gat is not better than other supervised baselines.
By applying centrality adjustment, \method achieves superior performance to both classes of baselines
regardless of the number of available known scores.

\begin{table*}[!htbp]
\tablefont
\centering
\caption{
Out-of-domain prediction results on real-world datasets. \method consistently outperforms all baselines.
Numbers after $ \pm $~symbol are standard deviation from 5-fold cross validation.
Best results are in bold, and second best results are underlined.
}
\begin{tabular}{ c | r | r | r | r | r | r }
\toprule
\multirow{2}{*}{\textbf{Method}} & \multicolumn{2}{c|}{\textbf{\music}} & \multicolumn{2}{c|}{\textbf{\tmdb}} & \multicolumn{2}{c}{\textbf{\imdb}} \\
& \makecell{\ndcg} & \makecell{\ndcgt} & \makecell{\ndcg} & \makecell{\ndcgt} & \makecell{\ndcg} & \makecell{\ndcgt} \\
\midrule
\pr & $ 0.6520 \pm 0.000 $ & $ 0.8779 \pm 0.000 $ & $ 0.8337 \pm 0.000 $ & $ 0.8079 \pm 0.000 $ & $ 0.0000 \pm 0.000 $ & $ 0.1599 \pm 0.000 $ \\
\ppr & $ \underline{0.7324 \pm 0.006} $ & $ \underline{0.9118 \pm 0.002} $ & $ 0.8060 \pm 0.041 $ & $ 0.7819 \pm 0.022 $ & $ 0.0000 \pm 0.000 $ & $ 0.1599 \pm 0.000 $ \\
\har & $ 0.7113 \pm 0.004 $ & $ 0.8982 \pm 0.001 $ & $ \underline{0.8913 \pm 0.010} $ & $ \underline{0.8563 \pm 0.007} $ & $ 0.2551 \pm 0.019 $ & $ 0.3272 \pm 0.005 $ \\ \hline
\linreg & $ 0.6644 \pm 0.006 $ & $ 0.8667 \pm 0.001 $ & $ 0.4990 \pm 0.013 $ & $ 0.5984 \pm 0.002 $ & $ 0.3064 \pm 0.007 $ & $ 0.2755 \pm 0.003 $ \\
\randforest & $ 0.6898 \pm 0.022 $ & $ 0.8796 \pm 0.003 $ & $ 0.5993 \pm 0.040 $ & $ 0.6236 \pm 0.005 $ & $ \underline{0.4066 \pm 0.145} $ & $ 0.3719 \pm 0.040 $ \\
\neuralnet & $ 0.6981 \pm 0.017 $ & $ 0.8836 \pm 0.005 $ & $ 0.5675 \pm 0.023 $ & $ 0.6172 \pm 0.009 $ & $ 0.2158 \pm 0.035 $ & $ 0.3105 \pm 0.019 $ \\
\gat & $ 0.6909 \pm 0.009 $ & $ 0.8834 \pm 0.003 $ & $ 0.5349 \pm 0.016 $ & $ 0.5999 \pm 0.007 $ & $ 0.3858 \pm 0.065 $ & $ \underline{0.4209 \pm 0.016} $ \\
\midrule
\textbf{\method} & $ \textbf{0.7964} \pm \textbf{0.007} $ & $ \textbf{0.9121} \pm \textbf{0.002} $ & $ \textbf{0.9078} \pm \textbf{0.004} $ & $ \textbf{0.8776} \pm \textbf{0.002} $ & $ \textbf{0.4519} \pm \textbf{0.051} $ & $ \textbf{0.4962} \pm \textbf{0.025} $ \\
\bottomrule
\end{tabular}
\label{tab:exp:realworld:outofdomain}
\end{table*}

\subsection{Analysis of \method}
\label{sec:exp:analysis}

\subsubsection{Effect of Considering Predicates}

To see how the consideration of predicates affects model performance,
we run \method on \fb, which has the largest number of predicates, and report \ndcg and \spearman
when a single embedding is used for all predicates (denoted by ``shared embedding'') vs. 
when each predicate uses its own embedding (denoted by ``distinct embedding'').
Note that using ``shared embedding'', \method loses the ability to distinguish between different predicates.
In the results given in \Cref{tab:exp:analysis:predicate}, we observe that 
\ndcg and \spearman are increased by 3.6\% and 12.7\%, respectively,
when a dedicated embedding is used for each predicate.
This shows that \method successfully makes use of predicates for modeling the relation between node importance;
this is especially crucial in KGs such as \fb that consist of a large number of predicates.

\begin{table}[!h]
\tablefont
\centering
\caption{Performance of \method on \fb when a single embedding is used for all predicates (shared embedding) vs. when each predicate uses its own embedding (distinct embedding).}
\begin{tabular}{ c | r | r }
\toprule
\makecell{\textbf{Metric}} & \makecell{\textbf{Shared Embedding}} & \makecell{\textbf{Distinct Embedding}} \\
\midrule
\makecell{\ndcg} & $ 0.9062 \pm 0.008 $ & $ \textbf{0.9385} \pm \textbf{0.004}$ \\
\makecell{\spearman} & $ 0.6894 \pm 0.007 $ & $\textbf{0.7772} \pm \textbf{0.006}$ \\
\bottomrule
\end{tabular}
\label{tab:exp:analysis:predicate}
\end{table}

\subsubsection{Flexibility for Centrality Adjustment.}

\begin{table}[!htbp]
\small
\centering
\setlength{\tabcolsep}{0.75mm}
\caption{Performance of \pr, log in-degree baseline, and \method with fixed and flexible centrality adjustment (CA) on \fb and \tmdb.}
\makebox[0.47\textwidth][c]{
\begin{tabular}{ c | r | r | r | r }
\toprule
\multirow{2}{*}{\textbf{\makecell{Method}}} & \multicolumn{2}{c|}{\textbf{\fb}} & \multicolumn{2}{c}{\textbf{\tmdb}} \\
& \makecell{\ndcg} & \makecell{\spearman} & \makecell{\ndcg} & \makecell{\spearman} \\
\midrule
\makecell{\pr} & $ 0.835 \pm 0.02 $ & $ 0.352 \pm 0.02 $ & $ 0.829 \pm 0.03 $ & $ 0.590 \pm 0.01 $ \\
\makecell{Log In-Degree} & $ 0.810 \pm 0.02 $ & $ 0.300 \pm 0.03 $ & $ 0.852 \pm 0.02 $ & $ 0.685 \pm 0.02 $ \\
\makecell{\method-Fixed CA} & $ 0.868 \pm 0.01 $ & $ 0.613 \pm 0.01 $ & $ 0.899 \pm 0.01 $ & $ 0.771 \pm 0.01 $ \\
\makecell{\textbf{\method-Flexible CA}} & $ \textbf{0.938} \pm \textbf{0.00} $ & $ \textbf{0.777} \pm \textbf{0.01} $ & $ \textbf{0.905} \pm \textbf{0.01} $ & $ \textbf{0.780} \pm \textbf{0.01} $ \\
\bottomrule
\end{tabular}
}
\vspace{-1em}
\label{tab:exp:analysis:flexiblecentrality}
\end{table}

In \Cref{eq:flexible_ca}, we perform scaling and shifting of $ c(i) $ for flexible centrality adjustment (CA).
Here we evaluate the model with fixed CA without scaling and shifting where the final estimation
$ s^*(i) = \sigma_s ( c(i) \cdot s^{L}(i) )$.
In \Cref{tab:exp:analysis:flexiblecentrality}, 
we report the performance of \method on \fb and \tmdb obtained with fixed and flexible CA while all other parameters were identical.
When node centrality strongly correlates with input scores, fixed CA obtains similar results to flexible CA.
This is reflected on the result of \tmdb dataset, where \pr and log in-degree baseline (\logindeg), 
which estimates node importance as the log of its in-degree, 
both estimate node importance close to the input scores.
On the other hand, when node centrality is not in good agreement with input scores, 
as demonstrated by the poor performance of \pr and LID as on \fb, 
flexible CA performs much better than fixed CA (8\% higher \ndcg, and 27\% higher \spearman on \fb).

\subsubsection{Parameter Sensitivity}
\begin{figure*}[!htbp]
\centering
\begin{subfigure}[t]{0.245\textwidth}
	\centering
	\includegraphics[width=.99\linewidth]{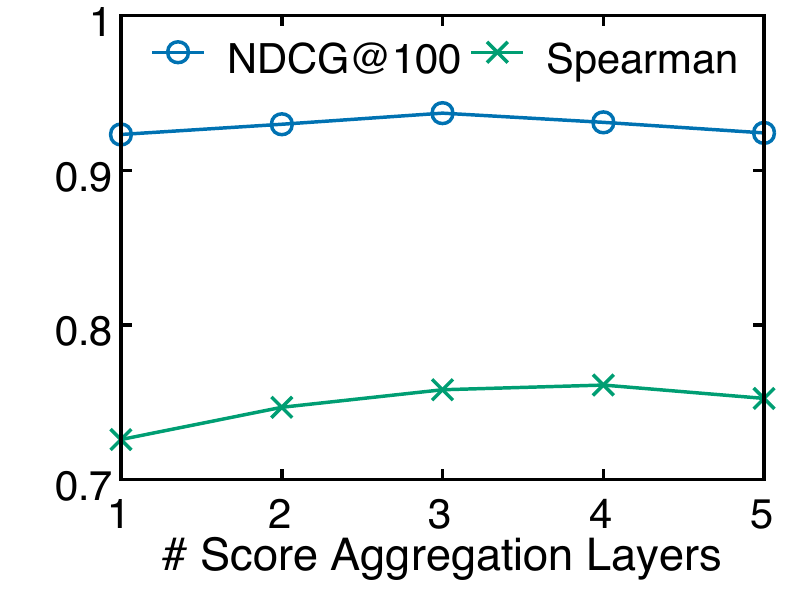}
\label{fig:sensitivity:num_layers}
\end{subfigure}
\begin{subfigure}[t]{0.245\textwidth}
	\centering
	\includegraphics[width=.99\linewidth]{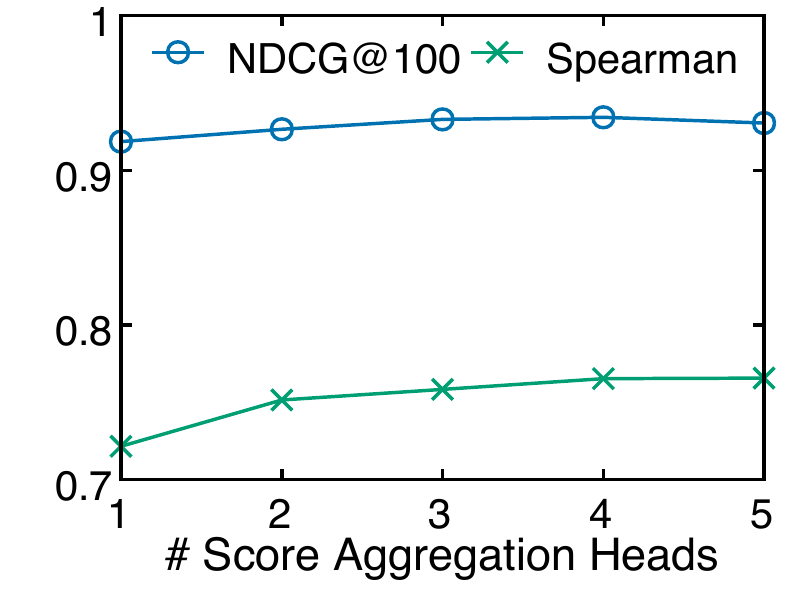}
\label{fig:sensitivity:num_heads}
\end{subfigure}
\begin{subfigure}[t]{0.245\textwidth}
	\centering
	\includegraphics[width=.99\linewidth]{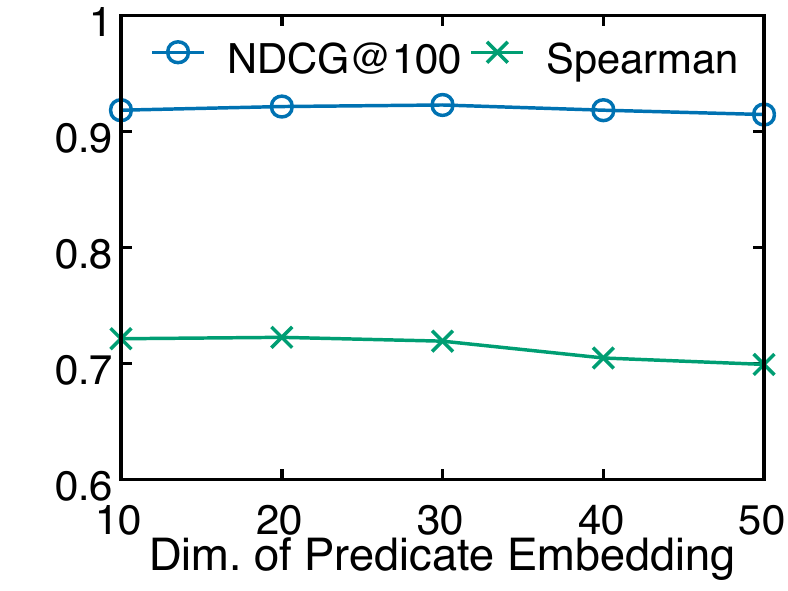}
\label{fig:sensitivity:dim_pred_embd}
\end{subfigure}
\begin{subfigure}[t]{0.245\textwidth}
	\centering
	\includegraphics[width=.99\linewidth]{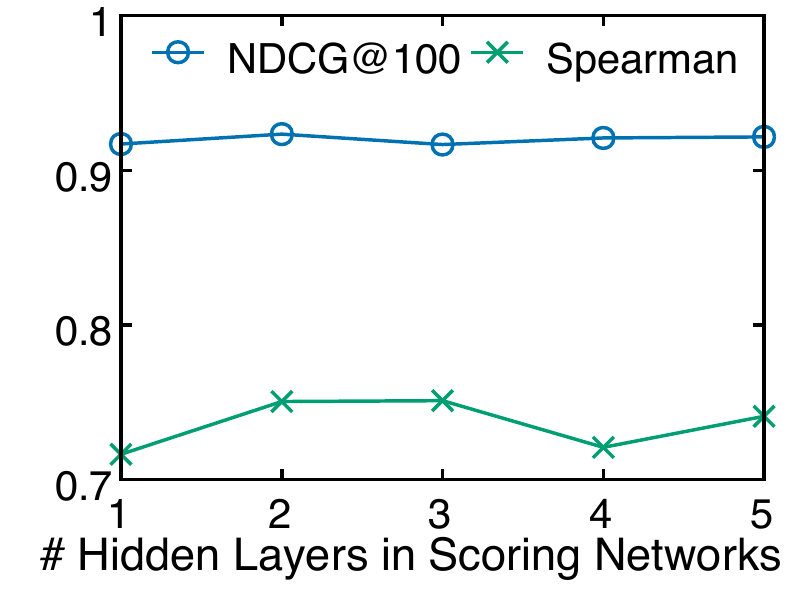}
\label{fig:sensitivity:scoring_num_layers}
\end{subfigure}
\vspace{-1.75\baselineskip}
\caption{Parameter sensitivity of \method on \fb. We report results varying one parameter on x-axis, while fixing all others.}
\label{fig:sensitivity}
\vspace{-1.70em}
\end{figure*}

We evaluate the parameter sensitivity of \method by measuring performance on \fb
varying one of the following parameters while fixing others to their default values (shown in parentheses):
number of score aggregation (SA) layers (1), number of SA heads in each SA layer (1), dimension of predicate embedding (10), and number of hidden layers in scoring networks (1 layer with 48 units).
Results presented in \Cref{fig:sensitivity}
shows that the model performance tends to improve as we use a greater number of SA layers and SA heads.
For example, \spearman increases from 0.72 to 0.77 as the number of SA heads is increased from 1 to 5.
Using more hidden layers for scoring networks also tends to boost performance, although exceptions are observed.
Increasing the dimension of predicate embedding beyond an appropriate value negatively affects the model performance,
although \method still achieves high \spearman compared to baselines.

\vspace{-1.5em}

\section{Related Work}
\label{sec:related}
\textbf{Node Importance Estimation.}
Many approaches have been developed for node importance estimation~\cite{page1999pagerank,DBLP:conf/www/Haveliwala02,DBLP:journals/kais/TongFP08,DBLP:conf/sigmod/JungPSK17,DBLP:conf/sdm/LiNY12,DBLP:journals/jacm/Kleinberg99}.
PageRank (\pr)~\cite{page1999pagerank} is based on the random surfer model
where an imaginary surfer randomly moves to a neighboring node with probability $ d $, 
or teleports to any other node randomly with probability $ 1 - d $.
\pr predicts the node importance to be the limiting probability of 
the random surfer being at each node.
Accordingly, \pr scores are determined only by the graph structure, and unaware of input importance scores.
Personalized PageRank (\ppr)~\cite{DBLP:conf/www/Haveliwala02} deals with
this limitation by biasing the random walk to teleport to a set of nodes
relevant to some specific topic, or alternatively, nodes with known importance scores.
Random walk with restart (\rwr)~\cite{DBLP:journals/kais/TongFP08,DBLP:conf/sigmod/JungPSK17} is a closely related method
that addresses a special case of \ppr where teleporting is restricted to a single node. \ppr and \rwr, however, are not well suited for KGs since they do not consider edge types.
To make a better use of rich information in KGs,
\har~\cite{DBLP:conf/sdm/LiNY12} extends the idea of random walk used by \pr and \ppr 
to solve limiting probabilities arising from multi-relational data, and 
distinguishes between different predicates in KGs while being aware of importance scores.
Previous methods can be categorized as non-trainable approaches 
with a fixed model structure that do not involve model parameter optimization.
In this paper, we explore supervised machine learning algorithms with a focus on graph neural networks.

\textbf{Graph Neural Networks (GNNs).}
GNNs are a class of neural networks that learn from arbitrarily structured graph data.
Many GNN formulations have been based on the notion of graph convolutions.
The pioneering work of Bruna et al.~\cite{DBLP:journals/corr/BrunaZSL13} defined the convolution operator in the Fourier domain,
which involved performing the eigendecomposition of the graph Laplacian;
as a result, its filters were not spatially localized, and computationally costly.
A number of works followed to address these limitations.
Henaff et al.~\cite{DBLP:journals/corr/HenaffBL15} introduced a localization of spectral filters via the spline parameterization.
Defferrard et al.~\cite{DBLP:conf/nips/DefferrardBV16} designed more efficient, strictly localized convolutional filters.
Kipf and Welling~\cite{DBLP:journals/corr/KipfW16} further simplified localized spectral convolutions via a first-order approximation.
To reduce the computational footprint and improve performance, 
recent works explored different ways of neighborhood aggregation.
One direction has been to restrict neighborhoods via sampling techniques 
such as uniform neighbor sampling~\cite{DBLP:conf/nips/HamiltonYL17}, vertex importance sampling~\cite{chen2018fastgcn}, 
and random walk-based neighbor importance sampling~\cite{DBLP:conf/kdd/YingHCEHL18}.
Graph attention networks (GAT)~\cite{gat2018}, which is most closely related to our method,
explores an orthogonal direction of assigning different importance to different neighbors 
by employing self-attention over neighbors~\cite{DBLP:conf/nips/VaswaniSPUJGKP17}.
While GAT exhibited state-of-the-art results, it was applied only to node classifications, and is unaware of predicates.
Building upon recent developments in GNNs, \method tackles the challenges for node importance estimation in KGs, 
which have not been addressed by existing GNNs.

\vspace{-1.5em}

\section{Conclusion}
\label{sec:concl}
Estimating node importance in KGs is an important problem with many applications 
such as item recommendation and resource allocation.
In this paper, we present a method \method that addresses this problem
by utilizing rich information available in KGs in a flexible manner
which is required to model complex relation between entities and their importance.
Our main ideas can be summarized as score aggregation via predicate-aware attention mechanism and flexible centrality adjustment.
Experimental results on predicting node importance in real-world KGs show that 
\method outperforms existing approaches, achieving 5--17\% higher NDCG@100 than the state of the art.
For future work, we will consider multiple independent input sources for node importance.
\vspace{-1em}

\bibliographystyle{ACM-Reference-Format}

\clearpage
\appendix
In the appendix, we provide details on datasets, experimental settings, and
additional experimental results, such as a case study on \tmdb and regression performance evaluation for in-domain predictions.

\abovedisplayskip=2.0pt
\belowdisplayskip=2.0pt

\begin{table*}[!h]
\vspace{-1em}
\caption{Top-10 movies and directors with highest predicted importance scores by \method, \har, and \gat on \tmdb. 
``\textit{ground truth rank}''$ - $``\textit{estimated rank}'' is shown for each prediction.}
\centering
\makebox[1.0\textwidth][c]{
\begin{subtable}{.5\textwidth}
\footnotesize
\caption{Top-10 movies (in-domain estimation).
A \textit{ground truth rank} is computed from known importance scores of movies used for testing.
}
\centering
\setlength{\tabcolsep}{0.75mm}
\begin{tabular}{ c | c | c | c | c | c | c }
	\toprule
	& \multicolumn{2}{c|}{\textbf{\method}} & \multicolumn{2}{c|}{\textbf{\har}} & \multicolumn{2}{c}{\textbf{\gat}} \\
	\midrule
	1 & The Dark Knight Rises & 11 & Jason Bourne & 63 & The Dark Knight Rises & 11 \\
	2 & The Lego Movie & 70 & The Wolf of Wall Street & 21 & Clash of the Titans & 103 \\
	3 & Spectre & 10 & Rock of Ages & 278 & Ant-Man & 4 \\
	4 & Les Misérables & 94 & Les Misérables & 94 & The Lego Movie & 68 \\
	5 & \makecell{The Amazing\\ Spider-Man} & 22 & The Dark Knight Rises & 7 & Jack the Giant Slayer & 126 \\
	6 & Toy Story 2 & 39 & V for Vendetta & 27 & Spectre & 7 \\
	7 & V for Vendetta & 26 & Now You See Me 2 & 81 & The Wolf of Wall Street & 16 \\
	8 & Clash of the Titans & 97 & Spectre & 5 & The 5th Wave & 67 \\
	9 & Ant-Man & -2 & \makecell{Austin Powers in \\Goldmember} & 140 & \makecell{The Hunger Games: \\Mockingjay - Part 2} & -4 \\
	10 & Iron Man 2 & 29 & Alexander & 141 & X-Men: First Class & 767 \\
	\bottomrule
\end{tabular}
\label{tab:exp:casestudy:indomain}
\end{subtable}
\quad\quad
\begin{subtable}{.5\textwidth}
\small
\caption{Top-10 directors (out-of-domain estimation).
A \textit{ground truth rank} corresponds to the rank in a director ranking (N/A indicates that the director is not in the director ranking).
}
\centering
\setlength{\tabcolsep}{0.75mm}
\begin{tabular}{ c | c | c | c | c | c | c }
	\toprule
	& \multicolumn{2}{c|}{\textbf{\method}} & \multicolumn{2}{c|}{\textbf{\har}} & \multicolumn{2}{c}{\textbf{\gat}} \\
	\midrule
	1 & Steven Spielberg & 0 & Steven Spielberg & 0 & Noam Murro & N/A \\
	2 & Tim Burton & 9 & Martin Scorsese & 44 & J Blakeson & N/A \\
	3 & Ridley Scott & 6 & Ridley Scott & 6 & Pitof & N/A \\
	4 & Martin Scorsese & 42 & Clint Eastwood & 19 & Paul Tibbitt & N/A \\
	5 & Francis Ford Coppola & 158 & Woody Allen & 112 & Rupert Sanders & N/A \\
	6 & Peter Jackson & -4 & Robert Zemeckis & 1 & Alan Taylor & 145 \\
	7 & Robert Rodriguez & 127 & Tim Burton & 4 & Peter Landesman & N/A \\
	8 & Gore Verbinski & 8 & David Fincher & 40 & Hideo Nakata & N/A \\
	9 & Joel Schumacher & 63 & Oliver Stone & 105 & Drew Goddard & N/A \\
	10 & Robert Zemeckis & -3 & Ron Howard & -2 & Tim Miller & N/A \\
	\bottomrule
\end{tabular}
\label{tab:exp:casestudy:outofdomain}
\end{subtable}
}
\vspace{-2em}
\label{tab:exp:casestudy}
\end{table*}

\section{Datasets}
\label{appendix:datasets}

We perform evaluation using four real-world KGs that have different characteristics.
All KGs were constructed from public data sources, which we specify in the footnote.
Summaries of these datasets (such as the number of nodes, edges, and predicates) are given in \Cref{tab:datasets}.
Below, we provide details on the construction of each KG.

\textbf{\fb.} We used a sample of Freebase\footnote{\url{https://everest.hds.utc.fr/doku.php?id=en:smemlj12}} 
used by \cite{DBLP:conf/nips/BordesUGWY13}.
The original dataset is divided into training, validation, and test sets.
We combined them into a single dataset, and later divided them randomly into three sets 
based on our proportion for training, validation, and test data.
In order to find the number of pageviews of a Wikipedia page, which is the importance score used for \fb,
we used Freebase/Wikidata mapping\footnote{\url{https://developers.google.com/freebase/}}.
Most entities in \fb can be mapped to the corresponding Wikidata page, from which we found the link to the item's English Wikipedia page,
which provides several information including the number of pageviews in the past 30 days.

\textbf{\music.} We built \music from the sample\footnote{\url{https://think.cs.vt.edu/corgis/csv/music/music.html}} of the Million Song Dataset\footnote{\url{https://labrosa.ee.columbia.edu/millionsong/}}.
This dataset is a collection of audio features and metadata for one million popular songs.
Among others, this dataset includes information about songs such as the primary artist and the album the song belongs to.
We constructed \music by adding nodes for these three entities (i.e., songs, artists, and albums), and 
edges of corresponding types between them as appropriate.
Note that \music is much more fragmented than other datasets.

\textbf{\tmdb.} We constructed \tmdb from the TMDb 5000 movie dataset\footnote{\url{https://www.kaggle.com/tmdb/tmdb-movie-metadata}}.
This dataset contains movies and relevant information such as movie genres, companies, countries, crews, and casts in a tabular form.
We added nodes for each of these entities, and added edges between two related entities with appropriate types.
For instance, given that ``Steven Spielberg'' directed ``Schindler's List'',
we added two corresponding director and movie nodes, and added an edge of type ``directed'' between them.

\textbf{\imdb.} We created \imdb from public IMDb datasets\footnote{\url{https://www.imdb.com/interfaces/}}.
IMDb datasets consist of several tables, which contain information such as titles, genres, directors, writers, principal casts and crews.
As for \tmdb, we added nodes for these entities, and connected them with edges of corresponding types.
In creating \imdb, we focused on entities related to movies, and excluded other entities that have no relation with movies.
In addition, IMDb datasets include titles each person is known for; 
we added edges between a person and these titles to represent this special relationship.

\textbf{Scores.}
For \fb, \tmdb, \imdb, we added 1 to the importance scores as an offset, and log-transformed them
as the scores were highly skewed.
For \music, two types of provided scores were all between 0 and 1, and we used them without log transformation.

\section{Experimental Settings}
\label{appendix:settings}

\subsection{Cross Validation and Early Stopping}

We performed 5-fold cross validation; i.e., for each fold, 80\% of the ground truth scores were used for training, and 
the other 20\% were used for testing.
For methods based on neural networks, we applied early stopping
by using 15\% of the original training data for validation and the remaining 85\% for training,
with a patience of 50.
That is, the training was stopped if the validation loss did not decrease for 50 consecutive epochs,
and the model with the best validation performance was used for testing.

\vspace{-1em}
\subsection{Software}

We used several open source libraries, and used Python~3.6 for our implementation.

\textbf{Graph Library.}
We used NetworkX 2.1 for graphs and graph algorithms:
\textit{MultiDiGraph} class was used for all KGs as there can be multiple edges of different types between two entities;
NetworkX's \textit{pagerank\_scipy} function was used for \pr and \ppr.

\textbf{Machine Learning Library.}
We chose TensorFlow 1.12 as our deep learning framework.
We used scikit-learn 0.20.0 for other machine learning algorithms such as random forest and linear regression.

\textbf{Other Libraries and Algorithms.}
For \gat, we used the reference TensorFlow implementation provided by the authors\footnote{\url{https://github.com/PetarV-/GAT}}.
We implemented \har in Python 3.6 based on the algorithm description presented in~\cite{DBLP:conf/sdm/LiNY12}.
For node2vec, we used the implementation available from the project page\footnote{\url{https://snap.stanford.edu/node2vec/}}.
NumPy 1.15 and SciPy 1.1.0 were used for data manipulation.

\vspace{-1em}
\subsection{Hyperparameters and Configurations}

\paragraph{\textbf{\textup{PageRank (\pr) and Personalized PageRank (\ppr)}}}
We used the default values for NetworkX's \textit{pagerank\_scipy} function with 0.85 as a damping factor.

\textbf{HAR}~\cite{DBLP:conf/sdm/LiNY12}.
As in \ppr, normalized input scores were used as probabilities for entities;
equal probability was assigned to all relations.
We set $ \alpha=0.15 $, $ \beta=0.15 $, $ \gamma=0 $.
The maximum number of iterations was set to 30.
Note that \har is designed to compute two types of importance scores, hub and authority. 
For \music, \tmdb, and \imdb KGs, these scores are identical 
since each edge in these graphs has a matching edge with an inverse predicate going in the opposite direction. 
Thus for these KGs, we only report authority scores. 
For \fb, we compute both types of scores, and report authority scores as hub scores are slightly worse overall.

\textbf{Linear Regression (\linreg) and Random Forests (\randforest)}.
For both methods, we used default parameter values defined by scikit-learn. 

\textbf{Neural Networks (\neuralnet)}.
Let $ [n_1, n_2, n_3, n_4] $ denote a 3-layer neural network 
where $ n_1, n_2, n_3 $ and $ n_4 $ are the number of neurons in the input, first hidden, second hidden, and output layers, respectively.
For \neuralnet, we used an architecture of [$ N_F, 0.5 \times N_F, 0.25 \times N_F, 1 $] where $ N_F $ is the dimension of node features.
We applied a rectified linear unit (ReLU) non-linearity at each layer, and 
used Adam optimizer with a learning rate $ \alpha = 0.001 $, $ \beta_1 = 0.9 $, $ \beta_2 = 0.999 $, and a weight decay of 0.0005.
\vspace{-0.25em}

\textbf{Graph Attention Networks (\gat)}~\cite{gat2018}.
We used a \gat model with two attentional layers, each of which consists of four attention heads, 
which is followed by a fully connected \neuralnet (\fcnn).
Following the settings in~\cite{gat2018},
we used a Leaky ReLU with a negative slope of 0.2 for attention coefficient computation, and
applied an exponential linear unit (ELU) non-linearity to the output of each attention head.
The output dimension of an attention head in all layers except the last was set to $ \max(0.25 \times N_F, 20) $.
For \fcnn after the attentional layers, we used an architecture of [$ 0.75 \times N_F $, $ 1 $] with ReLU as non-linearity.
Adam optimizer was applied with a learning rate $ \alpha = 0.005 $, $ \beta_1 = 0.9 $, $ \beta_2 = 0.999 $, and a weight decay of 0.0005.

\textbf{\method.}
We used an architecture where each score aggregation (SA) layer contains four SA heads.
For \fb, we used a model with three SA layers, and for other KGs, we used a model with one SA layer.
For \textsc{ScoringNetwork}, a two-layer \fcnn with an architecture of [$N_F, 0.75 \times N_F, 1 $] was used.
\method was trained with Adam optimizer using a learning rate $ \alpha=0.005 $, 
$ \beta_1 = 0.9 $, $ \beta_2 = 0.999 $, and a weight decay of 0.0005.
The dimension of predicate embedding was set to $ 10 $ for all KGs.
We used a Leaky ReLU with a negative slope of 0.2 for attention coefficient computation ($ \sigma_a $), and
a RELU for the final score estimation ($ \sigma_s $).
We defined $ \N(i) $ as outgoing neighbors of node $ i $.
Similar results were observed when we defined $ \N(i) $ to include both outgoing and incoming neighbors of node~$ i $.
Since the initial values for $ \gamma $ and $ \beta $ (parameters for centrality adjustment) affect model performance, 
we determined these initial values for each dataset based on the validation performance.

\textbf{node2vec}~\cite{DBLP:conf/kdd/GroverL16}.
We set the number of output dimensions to 64 for \fb, \music, and \tmdb, and 128 for \imdb.
Other parameters were left to their default values.
Note that node2vec was used in our experiments to generate node features for supervised methods.

\vspace{-2em}
\section{Additional Evaluation}

\subsection{Case Study}
\label{appendix:casestudy}

We take a look at the predictions made by \method, \har, and \gat on \tmdb.
Given popularity scores for some movies, methods estimate the importance score of all other entities in \tmdb.
Among them, \Cref{tab:exp:casestudy} reports the top-10 movies and directors that are estimated to have the highest importance scores by each method
with ``ground truth rank''$-$``estimated rank'' shown for each entity.

\textbf{In-domain estimation} is presented in \Cref{tab:exp:casestudy:indomain}.
A ground truth rank is computed from the known importance scores of movies reserved for testing.
The top-10 movies predicted by \method is qualitatively better than the two others.
For example, among the ten predictions of \gat and \har, 
the difference between ground truth rank and predicted rank is greater than 100 for three movies.
On the other hand, the rank difference for \method is less than 100 for all predictions.

\textbf{Out-of-domain estimation} is presented in \Cref{tab:exp:casestudy:outofdomain}.
As importance scores for directors are unknown, 
we use the director ranking introduced in \Cref{sec:exp:datasets}.
A ground truth rank denotes the rank in the director ranking, and ``N/A'' indicates that 
the director is not included in the director ranking.
The quality of the top-10 directors estimated by \method and \har is similar to each other
with five directors appearing in both rankings (e.g., Steven Spielberg).
Although \gat is not considerably worse than \method for in-domain estimation,
its out-of-domain estimation is significantly worse than others:
nine out of ten predictions are not even included in the list of top-200 highest earning directors.
By respecting node centrality, \method yields a much better ranking consistent with ground truth.

\vspace{-1em}
\subsection{Regression Performance Evaluation for In-Domain Predictions}
\label{appendix:regression_performance}

In order to see how accurately supervised approaches recover the importance of nodes, 
we measure the regression performance of their in-domain predictions.
In particular, we report RMSE (root-mean-squared error) of supervised methods in \Cref{tab:exp:realworld:indomain:rmse}.
Non-trainable methods are excluded since their output is not in the same scale as the input scores.
\method performs better than other supervised methods on all four real-world datasets.
Overall, the regression performance of supervised approaches follows a similar trend to 
their performance in terms of ranking measures reported in \Cref{tab:exp:realworld:indomain}.
\vspace{-1em}

\begin{table}[!htbp]
\tablefont
\centering
\setlength{\tabcolsep}{0.5mm}
\caption{
	RMSE (root-mean-squared error) of in-domain prediction for supervised methods. 
	Lower RMSE is better.
	\method consistently outperforms all baselines.
	Numbers after $ \pm $~symbol are standard deviation from 5-fold cross validation.
	Best results are in bold, and second best results are underlined.
}
\makebox[0.4\textwidth][c]{
	\begin{tabular}{ c | r | r | r | r}
		\toprule
		\textbf{Method} & \multicolumn{1}{c|}{\textbf{\fb}} & \multicolumn{1}{c|}{\textbf{\music}} & \multicolumn{1}{c|}{\textbf{\tmdb}} & \multicolumn{1}{c}{\textbf{\imdb}} \\
		\midrule
		\linreg & $ 1.3536 \pm 0.017 $ & $ 0.1599 \pm 0.002 $ & $ 0.8431 \pm 0.028 $ & $ 1.7534 \pm 0.005 $ \\
		\randforest & $ 1.2999 \pm 0.024 $ & $ \underline{0.1494 \pm 0.002} $ & $ 0.9223 \pm 0.015 $ & $ 1.8181 \pm 0.011 $ \\
		\neuralnet & $ 1.2463 \pm 0.015 $ & $ 0.1622 \pm 0.009 $ & $ 0.8496 \pm 0.012 $ & $ 2.0279 \pm 0.033 $ \\
		\gat & $ \underline{1.0798 \pm 0.031} $ & $ 0.1635 \pm 0.007 $ & $ \underline{0.8020 \pm 0.010} $ & $ \underline{1.2972 \pm 0.018} $ \\
		\midrule
		\textbf{\method} & $ \textbf{0.9471} \pm \textbf{0.017} $ & $ \textbf{0.1491} \pm \textbf{0.002} $ & $ \textbf{0.7150} \pm \textbf{0.003} $ & $ \textbf{1.2079} \pm \textbf{0.011} $ \\
		\bottomrule
	\end{tabular}
}
\label{tab:exp:realworld:indomain:rmse}
\end{table}

\end{document}